\pdfoutput=1

\documentclass[11pt]{article}
\usepackage[preprint]{acl}

\usepackage{times}
\usepackage{latexsym}
\usepackage{multirow, makecell}

\usepackage[T1]{fontenc}
\usepackage[utf8]{inputenc}
\usepackage{amsfonts, amsmath, amssymb}
\usepackage{tikz-dependency}
\usepackage{microtype}
\usepackage{inconsolata}
\usepackage[usenames,dvipsnames]{xcolor}
\usepackage{subcaption}
\usepackage{graphicx}

\title{Hierarchical Bracketing Encodings Work for Dependency Graphs}
\author
{
    Ana Ezquerro, Carlos Gómez-Rodríguez and David Vilares\\
    Universidade da Coru\~{n}a, CITIC \\
    Departamento de Ciencias de la Computación y Tecnologías de la Información \\
    Campus de Elvi\~{n}a s/n, 15071 \\ A Coru\~{n}a, Spain \\
    \texttt{\{ana.ezquerro, carlos.gomez, david.vilares\}@udc.es} \\
}

\usepackage{contour}

\newcommand{\olb}{\texttt{<}}
\newcommand{\clb}{\texttt{\textbackslash}}
\newcommand{\orb}{\texttt{/}}
\newcommand{\crb}{\texttt{>}}

\definecolor{rred}{HTML}{CC1B00}
\newcommand{\ols}{\contour{black}{\texttt{<}}}
\newcommand{\cls}{\contour{black}{\texttt{\textbackslash}}}
\newcommand{\ors}{\contour{black}{\texttt{/}}}
\newcommand{\crs}{\contour{black}{\texttt{>}}}
\newcommand{\planeii}[1]{\textcolor{rred}{#1\textsuperscript{*}}}

\definecolor{b3color}{HTML}{3366CC}
\definecolor{b63color}{HTML}{DC3912}
\definecolor{b64color}{HTML}{109618}
\definecolor{hbcolor}{HTML}{990099}

\begin{document}
\maketitle
\begin{abstract}
We revisit hierarchical bracketing encodings from a practical perspective in the context of dependency graph parsing. The approach encodes graphs as sequences, enabling linear-time parsing with $n$ tagging actions, and still representing reentrancies, cycles, and empty nodes. Compared to existing graph linearizations, this representation substantially reduces the label space while preserving structural information. We evaluate it on a multilingual and multi-formalism benchmark, showing competitive results and consistent improvements over other methods in exact match accuracy. 
\end{abstract}

\section{Introduction}

Sequence labeling (SL) offers a simple yet effective paradigm for a wide range of natural language problems. By assigning one label to each token, sequence labeling models eliminate the need for complex decoding algorithms, and make inference simpler, faster and scalable in structured prediction settings. When paired with neural encoders \cite{ma-hovy-2016-end}---especially pre-trained encoders \cite{devlin-etal-2019-bert}---SL models achieve strong results with minimal architectural complexity. 

Recently, sequence labeling has been used to address various flavors of syntactic \emph{tree} parsing such as continuous \cite{gomez-rodriguez-vilares-2018-constituent,kitaev-klein-2020-tetra,amini-cotterell-2022-parsing} and discontinuous constituency parsing \cite{vilares-gomez-rodriguez-2020-discontinuous}, and dependency parsing \cite{strzyz-etal-2019-viable,vacareanu-etal-2020-parsing,amini-etal-2023-hexatagging,gomez-rodriguez-etal-2023-4}. Some approaches rely on positional offsets to indicate explicit relations between tokens, others use explicit bracketing schemes, and some derive from transition-based parsing systems. Despite this progress, extending SL to graph parsing introduces new challenges. Graphs may include reentrancies, cycles, and disconnected components---structures that cannot be directly captured by most tree-based linearizations. As a result, most graph parsers rely on graph- \cite{dozat-manning-2018-simpler, wang-etal-2019-second} or transition-based \cite{fernandez-gonzalez-gomez-rodriguez-2020-transition} decoders. To date, only \citet{ezquerro-etal-2024-dependency} have shown that SL can be effectively adapted to this setting, suggesting it may offer a viable alternative that simplifies decoding while keeping the linearizations learnable.

Our work builds upon recent efforts to encode edge information through bracketing schemes, previously applied to  dependency trees \cite{strzyz-etal-2019-viable} and graphs \cite{ezquerro-etal-2024-dependency}. Namely, we apply the optimal hierarchical bracketing encoding from \citet{ezquerro-etal-2025-hierarchical}, a framework theoretically defined for both trees and graphs but empirically validated only on dependency trees, and provide the first empirical evaluation on dependency graphs. We evaluate its non-projective variant on a large benchmark of graph annotations, where re-entrancies and cycles are possible. Experiments show that this encoding preserves the performance of the original bracketing encodings of~\citet{ezquerro-etal-2024-dependency}, while achieving a reduced label space and maintaining full coverage of dependency graphs. Notably, it improves exact match scores, likely due to a more balanced and compact label distribution.

\section{Bracketing encodings for graphs}

Let $W=(w_1,...,w_n)\in\mathcal{V}^n$  be an input sentence from a vocabulary $\mathcal{V}$. A dependency graph built upon $W$ is defined as $G=(W,A)$, where each element of $A$ is an arc $(h\to d)$, such that $h\in[0,n]$ and $d\in[1,n]$, that connects a dependent ($w_d$) with its head ($w_h$). Note that, unlike in dependency trees, $A$ is not constrained by connectivity and acyclicity properties. Following \citet{ezquerro-etal-2024-dependency}, we define an SL framework as an encoding ($\varepsilon:\mathcal{A}^n\to\mathcal{L}^n$)\footnote{Here we use $\mathcal{A}^n$ to denote all the possible sets of arcs for a graph of size $n$.} and decoding ($\delta:\mathcal{L}^n\to\mathcal{A}^n$) function that represent a graph’s arc information as a label sequence of length $n$ over a label set $\mathcal{L}$.

\begin{figure}[tbp]
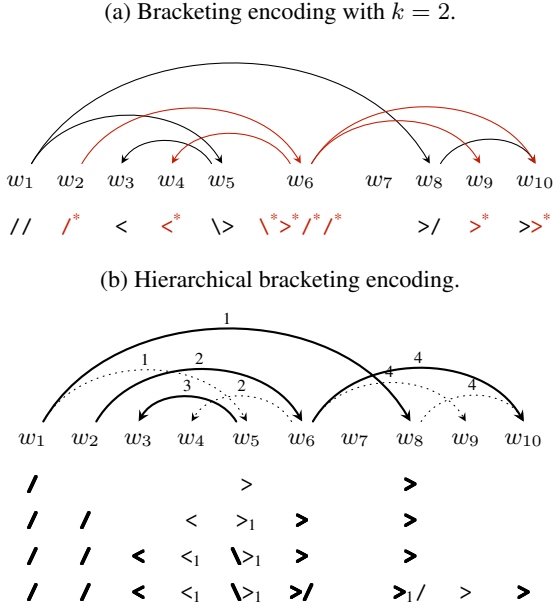
\centering\small
    \begin{subfigure}[t]{\linewidth}\centering\small
        \caption{\label{fig:bracket-example}Bracketing encoding with $k=2$.}
        \begin{dependency}[arc edge, text only label, hide label, label style={above}]
            \begin{deptext}[column sep=4pt]
                $w_1$ \& $w_2$ \& $w_3$ \& $ w_4$ \& $w_5$ \& $w_6$ \& $w_7$ \& $w_8$ \& $w_9$ \& $w_{10}$\\[1em]
                \orb\orb \& \planeii{\orb} \& \olb \& \planeii{\olb} \& \clb\crb\& \planeii{\clb}\planeii{\crb}\planeii{\orb}\planeii{\orb} \&  \&\crb\orb  \& \planeii{\crb} \& \crb\planeii{\crb} \\[0.5em]
            \end{deptext}
            % first plane
            \depedge{1}{5}{1}
            \depedge{1}{8}{2}
            \depedge{5}{3}{4}
            \depedge{8}{10}{6}
    
            % second plane
            \depedge[edge style={rred}]{2}{6}{3}
            \depedge[edge style={rred}]{6}{4}{5}
            \depedge[edge style={rred}]{6}{10}{7}
            \depedge[edge style={rred}]{6}{9}{8}
        \end{dependency}
    \end{subfigure}
    \begin{subfigure}[t]{\linewidth}\centering\small
        \caption{\label{fig:hierarchical-example}Hierarchical bracketing encoding.}
        \begin{dependency}[arc edge, text only label, label style={above}]
            \begin{deptext}[column sep=5pt]
                $w_1$ \& $w_2$ \& $w_3$ \& $ w_4$ \& $w_5$ \& $w_6$ \& $w_7$ \& $w_8$ \& $w_9$ \& $w_{10}$\\[1em]
                \ors \&  \& \& \& \crb \&  \& \& \crs \\[0.5em]
                \ors \& \ors  \& \& \olb \& \crb\textsubscript{1} \&\crs \&\& \crs \\[0.5em]
                \ors \& \ors  \& \ols \& \olb\textsubscript{1} \& \cls\crb\textsubscript{1} \& \crs \&\& \crs \\[0.5em]
                \ors \& \ors  \& \ols \& \olb\textsubscript{1} \& \cls\crb\textsubscript{1} \& \crs\ors \&\& \crs\textsubscript{1}\orb \& \crb\&  \crs \\[0.5em]
            \end{deptext}
            % first rope 
            \depedge[edge style={thick}]{1}{8}{1}
            \depedge[edge style={dotted}]{1}{5}{1}
    
            % second rope 
            \depedge[edge style={thick}]{2}{6}{2}
            \depedge[edge style={dotted}]{6}{4}{2}
    
            % third rope 
            \depedge[edge style={thick}]{5}{3}{3}
    
            % fourth rope c
            \depedge[edge style={thick}]{6}{10}{4}
            \depedge[edge style={dotted}]{6}{9}{4}
            \depedge[edge style={dotted}]{8}{10}{4}
        \end{dependency}
    \end{subfigure}
    \caption{Bracketing and hierarchical bracketing encoding for the same dependency graph. In Figure \ref{fig:bracket-example}, crossing arcs in the same direction and their associated brackets are colored in red. In Figure \ref{fig:hierarchical-example}, structural arcs of the rope cover and their symbols are shown in bold, while auxiliary arcs are in dotted lines. Arc labels identify the structural  set they belong to, while bracket subindices are used by the encoding to support crossing arcs. Each row shows the brackets added when encoding each structural set.}
\end{figure}

\citet{ezquerro-etal-2024-dependency} adapted \citet{strzyz-etal-2019-viable}'s bracketing encoding for dependency trees to graphs. In the bracketing encoding, each arc is represented with two balanced brackets that are included in the labels of its head and dependent: (\olb,\clb) for left arcs and (\orb,\crb) for right arcs. The decoding process uses a left-to-right pass that reads the sequence of brackets, pushing positions with opening brackets into each stack and popping elements when a closing bracket is found.  To solve the limitation of crossing arcs in the same direction (see Figure \ref{fig:bracket-example}), they proposed distributing the arcs of $A$ in $k$ \emph{relaxed planes}, where each plane does not contain crossing arcs in the same direction; and encode each plane with different bracket symbols, so only brackets from the same plane match each other at decoding time.

Although \citet{ezquerro-etal-2024-dependency} achieved strong performance on a large multilingual benchmark using $k \leq 3$, the bracketing encoding presents key limitations in terms of its theoretical coverage\footnote{Theoretical coverage is defined as the ratio of graphs in a reference treebank for which the linearization algorithm can produce a lossless encoding–decoding cycle, i.e., the original graph can be perfectly reconstructed from its encoded form.}. The hyperparameter $k$ inherently constrains the representation to graph structures with no more than $k$ relaxed planes. While \citet{ezquerro-etal-2024-dependency} explored increasing this value, doing so comes at the computational cost of expanding the label set $\mathcal{L}$. Their bounded bit-based encodings ($4k$ and $6k$-bit) address this issue and fix the cardinality of the label set, but they still rely on a fixed $k$, thereby maintaining the same limitation on theoretical coverage.

We fill this gap by operationalizing the concept of rope covers \cite{yli-jyra-2019-transition} within the hierarchical bracketing encoding, defined theoretically in \citet{ezquerro-etal-2025-hierarchical}, for graphs, to 
successfully apply a more compressed bracketing representation that removes the need for an hyperparameter $k$.

\section{Hierarchical Bracketing Encoding}
Given a graph $G=(W,A)$, a \emph{rope cover} is a subset $R\subseteq A$ such that every arc in $A\setminus R$ leans\footnote{We say that an arc $(h\to d)$ leans on another $(h'\to d')$ if  $(h'\to d')$  covers  $(h\to d)$ and either $\min(h,d) = \min(h',d')$ or $\max(h,d) = \max(h',d')$.} on at least one arc of $R$; and it is said to be \emph{proper} if no arc in $R$ leans on another arc of $R$. The arcs of $R$ as denoted as \emph{structural} arcs, and the arcs of $A\setminus R$ as \emph{auxiliary} arcs. \citet{yli-jyra-2019-transition} demonstrated that the proper rope cover is unique and defined an algorithm to find it for any arbitrary graph; and \citet{ezquerro-etal-2025-hierarchical} showed that it is optimal in the number of structural arcs. Extending this terminology, a proper rope cover $R$ gives rise to the concept of a structural set: a maximal subset $S \subseteq A$ that contains one arc from $R$, while all other arcs in $S$ lean on it.

Figure \ref{fig:hierarchical-example} shows the proper rope cover for a dependency graph. See that the first structural arc is $(1\to 8)$ and the only arc that leans on it is $(1\to 5)$, so these two arcs together form a structural set. The arc $(6\to 10)$ is also a structural arc and its auxiliary arcs are $(6\to 9)$ and $(8\to 10)$, and the set of these three arcs is also a structural set.

\paragraph{Encoding and decoding structural sets} \citet{ezquerro-etal-2025-hierarchical} proposed an algorithm to independently encode each structural set in an optimal number of unique labels, using \emph{balanced superbrackets} for structural arcs (\cls, \crs, \ols, \ors) and one bracket symbol (\clb, \crb, \olb, \orb) to encode the non-leant position of each auxiliary arc. See Figure \ref{fig:hierarchical-example} (row 1): the first structural set is $S_1=\{(1\to 8), (1\to 5)\}$ and $(1\to 8)$ is the structural arc, which is encoded with balanced superbrackets $(\ors,\crs)$. The auxiliary arc $(1\to 5)$ is encoded with only one bracket (\crb) on the non-leant position (in this case, 1 is the leant position, so the bracket corresponding to position 5 is the one encoded).

The decoding process reads the bracket sequence from left to right, pushing any opening symbol (\ors, \orb, \ols, \olb) into a stack. When a closing superbracket is found (\cls, \crs), the system pops stack elements, matching brackets with said superbracket, until an opening superbracket is found (\ols, \ors). Instead, when a closing bracket is found (\clb, \crb), the system matches the opening superbracket that should be on top of the stack with the closing bracket. 

\paragraph{Encoding and decoding crossing arcs} When jointly encoding different structural sets with crossing arcs, the decoding previously defined leads to incorrect arcs\footnote{Note that auxiliary arcs from the same structural set cannot cross each other in the same direction; and when they cross in different directions, they can still be recovered since the bracket direction already determines the superbracket that matches each arc.}. For instance, if we consider only the symbols in Figure \ref{fig:hierarchical-example} (row 2): (\ors, \ors, $\cdot$, \olb, \crb, \crs, $\cdot$, \crs, $\cdot$, $\cdot$), the decoding algorithm recovers $(2\to 5)$ instead of $(1\to 5)$ since the opening superbracket (\ors) of the structural arc $(2\to 6)$ is found first in the decoding stack. To solve this issue, we adopt the indexing approach from \citet{ezquerro-etal-2025-hierarchical} and add an index in those symbols that require skipping matching structural arcs at decoding time. In Figure \ref{fig:hierarchical-example} (row 2), when adding the index 1 to $w_5$'s bracket (\crb\textsubscript{1}), the decoding system skips one structural arc when parsing the stack, which results in matching the bracket \crb\textsubscript{1} with the structural arc in position 1, resolving the auxiliary arc $(1\to 5)$. When adding the structural arc $(5\to 3)$, $w_4$'s bracket index needs to be increased so it resolves to $(6\to 4)$ rather than $(5\to 4)$.

\paragraph{Postprocessing} As is common in parsing as sequence labeling, the mapping from graphs to label sequences is not surjective. Thus, when implementing this approach in practice by training a sequence labeling system to produce label sequences, the predicted sequences are not guaranteed to be structurally valid. To prevent decoding errors, such as popping an element from an empty stack, we apply a separate postprocessing step to correct the sequence and ensure a well-formed dependency graph. Specifically, this process matches any unbalanced closing brackets with the dummy start node ($w_0$) and discards any unclosed superbrackets left on the stack at the end of decoding.

\section{Experiments}
The main goal of this work is to assess whether hierarchical bracketing encodings can effectively support dependency graph parsing within a neural sequence labeling framework. Our source code to reproduce our experiments is available at \url{https://github.com/anaezquerro/separ}.

\begin{table*}[t!]\centering\small 
    \setlength{\tabcolsep}{3pt}
    \renewcommand{\arraystretch}{1.2}
    \begin{tabular}{c|ll|ll|ll|ll|ll|ll|}
        \cline{2-13}
        & \multicolumn{2}{c|}{\textbf{B\textsubscript{2}}} & \multicolumn{2}{c|}{\textbf{B\textsubscript{3}}} & \multicolumn{2}{c|}{\textbf{B6\textsubscript{3}}} & \multicolumn{2}{c|}{\textbf{B6\textsubscript{4}}} & \multicolumn{2}{c|}{\textbf{HB}} & \multicolumn{2}{c|}{\textbf{Biaf.}} \\
        \cline{2-13}
        & \makecell[c]{LF} & \makecell[c]{LM} & \makecell[c]{LF} & \makecell[c]{LM} & \makecell[c]{LF} & \makecell[c]{LM} & \makecell[c]{LF} & \makecell[c]{LM} & \makecell[c]{LF} & \makecell[c]{LM}& \makecell[c]{LF} & \makecell[c]{LM} \\
        \hline 
        \textit{en}\textsubscript{DM} & 94.57\textsubscript{100} & 52.27\textsubscript{100} & 94.67\textsubscript{100} & 52.48\textsubscript{100} & 94.04\textsubscript{99.51} & 43.69\textsubscript{83.62} & \textbf{94.74}\textsubscript{99.96} & 51.77\textsubscript{98.51} & 94.52 & \underline{\textbf{53.05}} & \underline{95.45} & 46.95 \\
        \textit{en}\textsubscript{PAS} & 95.27\textsubscript{*100} & 47.02\textsubscript{99.93} & 95.50\textsubscript{100} & 49.65\textsubscript{100} & 93.56\textsubscript{97.56} & 24.47\textsubscript{42.55} & 95.01\textsubscript{99.37} & 42.27\textsubscript{78.30} & \textbf{95.59} & \underline{\textbf{51.84}} & \underline{96.13} & 48.51 \\
        \textit{en}\textsubscript{PSD} & 85.33\textsubscript{99.96} & 15.82\textsubscript{98.58} & 85.95\textsubscript{*100} & 17.09\textsubscript{99.86} & 85.68\textsubscript{99.98} & 15.67\textsubscript{98.94} & \textbf{86.20}\textsubscript{*100} & \underline{\textbf{17.66}}\textsubscript{99.93} & 85.61 & 16.67 & \underline{86.84} & 16.60 \\
        \textit{cs}\textsubscript{PSD} & 90.02\textsubscript{99.96} & 28.80\textsubscript{98.02} & 90.03\textsubscript{100} & 29.40\textsubscript{100} & 90.25\textsubscript{99.98} & 29.52\textsubscript{99.22} & \textbf{90.48}\textsubscript{*100} & \underline{\textbf{31.56}}\textsubscript{99.82} & 89.64 & 30.84 & \underline{91.21} & 27.90 \\
        \textit{zh}\textsubscript{PAS} & 87.20\textsubscript{*100} & 31.72\textsubscript{99.77} & \textbf{88.67}\textsubscript{100} & 32.70\textsubscript{100} & 84.67\textsubscript{96.57} & 21.41\textsubscript{46.79} & 87.72\textsubscript{98.30} & 27.18\textsubscript{71.40} & 88.09 & \underline{\textbf{34.35}} & \underline{90.38} & 34.06 \\
        \hline 
        \textit{ar}\textsubscript{PADT} & 83.13\textsubscript{99.97} & 11.47\textsubscript{98.09} & 82.75\textsubscript{*100} & 10.15\textsubscript{99.85} & 82.72\textsubscript{99.97} & 11.62\textsubscript{98.68} & \textbf{83.17}\textsubscript{99.98} & \underline{\textbf{13.24}}\textsubscript{99.56} & 82.54 & 11.47 & \underline{85.26} & 11.03 \\
        \textit{bg}\textsubscript{BTB} & 93.97\textsubscript{100} & 47.76\textsubscript{100} & 92.89\textsubscript{100} & 47.94\textsubscript{100} & \textbf{94.29}\textsubscript{99.99} & \underline{\textbf{51.52}}\textsubscript{99.64} & 93.32\textsubscript{99.99} & 50.45\textsubscript{99.73} & 92.83 & 50.18 & \underline{94.92} & 48.75 \\
        \textit{fi}\textsubscript{TDT} & 90.35\textsubscript{99.93} & 42.57\textsubscript{97.94} & 90.50\textsubscript{*100} & 43.47\textsubscript{99.87} & \textbf{91.03}\textsubscript{99.93} & \underline{\textbf{46.24}}\textsubscript{98.39} & 90.81\textsubscript{99.97} & 45.47\textsubscript{99.16} & 90.10 & 45.59 & \underline{92.33} & 45.79 \\
        \textit{fr}\textsubscript{SEQ.} & 92.98\textsubscript{*100} & 37.72\textsubscript{99.56} & 92.47\textsubscript{100} & 38.16\textsubscript{100} & \textbf{93.98}\textsubscript{99.97} & 45.18\textsubscript{98.46} & 93.88\textsubscript{100} & \underline{\textbf{47.81}}\textsubscript{100} & 92.97 & 44.96 & \underline{94.91} & 47.59 \\
        \textit{it}\textsubscript{ISDT} & 93.16\textsubscript{*100} & 45.44\textsubscript{99.79} & 93.47\textsubscript{100} & 45.02\textsubscript{100} & 93.51\textsubscript{*100} & 47.93\textsubscript{99.59} & 93.52\textsubscript{100} & 48.76\textsubscript{100} & \textbf{93.75} & \underline{\textbf{49.38}} & \underline{94.36} & 48.13 \\
        \textit{lt}\textsubscript{ALK.} & 83.50\textsubscript{99.92} & 18.57\textsubscript{97.22} & 80.79\textsubscript{99.99} & 19.74\textsubscript{99.56} & \underline{\textbf{84.85}}\textsubscript{99.97} & 21.93\textsubscript{98.68} & 81.75\textsubscript{99.99} & \underline{\textbf{22.37}}\textsubscript{99.71} & 79.12 & 20.32 & 83.82 & 20.47 \\
        \textit{lv}\textsubscript{LVTB} & 87.80\textsubscript{99.95} & 37.47\textsubscript{98.41} & 87.12\textsubscript{99.99} & 35.27\textsubscript{99.78} & 87.78\textsubscript{99.94} & 38.29\textsubscript{98.03} & \textbf{88.30}\textsubscript{99.98} & \underline{\textbf{39.99}}\textsubscript{99.23} & 87.46 & 38.51 & \underline{89.84} & 39.66 \\
        \textit{pl}\textsubscript{PDB} & 93.25\textsubscript{99.97} & 51.33\textsubscript{98.69} & 93.61\textsubscript{*100} & 53.41\textsubscript{99.82} & \textbf{93.63}\textsubscript{99.97} & \underline{\textbf{53.72}}\textsubscript{98.92} & 93.61\textsubscript{99.99} & 53.27\textsubscript{99.50} & 92.60 & 52.46 & \underline{94.40} & 52.55 \\
        \textit{ru}\textsubscript{SYN.} & 93.74\textsubscript{99.99} & 53.12\textsubscript{99.65} & 93.64\textsubscript{100} & 52.47\textsubscript{100} & 93.86\textsubscript{*100} & 54.01\textsubscript{99.85} & \textbf{93.93}\textsubscript{*100} & 54.00\textsubscript{99.95} & 93.54 & \underline{\textbf{54.24}} & \underline{94.45} & 51.93 \\
        \textit{sk}\textsubscript{SNK} & 92.08\textsubscript{99.99} & 50.90\textsubscript{99.72} & 91.83\textsubscript{100} & 49.29\textsubscript{100} & \textbf{92.51}\textsubscript{99.99} & 52.87\textsubscript{99.62} & 92.41\textsubscript{*100} & 53.16\textsubscript{99.91} & 92.11 & \underline{\textbf{54.85}} & \underline{93.99} & 54.38 \\
        \textit{sv}\textsubscript{TAL.} & 89.90\textsubscript{*100} & 38.64\textsubscript{99.84} & 89.50\textsubscript{100} & 38.23\textsubscript{100} & \textbf{91.12}\textsubscript{99.98} & \underline{\textbf{45.28}}\textsubscript{99.34} & 90.97\textsubscript{*100} & 43.23\textsubscript{99.92} & 89.87 & 42.00 & \underline{92.05} & 41.43 \\
        \textit{ta}\textsubscript{TTB} & 61.89\textsubscript{100} & 1.67\textsubscript{100} & 61.89\textsubscript{100} & 1.67\textsubscript{100} & \underline{\textbf{66.08}}\textsubscript{100} & 2.50\textsubscript{100} & 66.08\textsubscript{100} & 2.50\textsubscript{100} & 63.67 & \underline{\textbf{3.33}} & 64.83 & 2.50 \\
        \textit{uk}\textsubscript{IU} & 89.94\textsubscript{99.99} & 34.30\textsubscript{99.10} & 89.96\textsubscript{100} & 32.74\textsubscript{100} & 90.38\textsubscript{99.99} & 37.33\textsubscript{99.10} & \textbf{90.59}\textsubscript{*100} & \textbf{37.89}\textsubscript{99.78} & 89.50 & 37.44 & \underline{92.17} & \underline{38.34} \\
        \hline 
        $\mu$ & 88.85 & 36.13 & 88.73 & 36.34 & 89.19 & 36.24 & \textbf{89.30} & 38.31 & 88.59 & \underline{\textbf{38.50}} & \underline{90.45} & 37.69 \\
        \hline 
    \end{tabular}
    \caption{\label{tab:results}LF and LM performance for the bracketing (\textbf{B}), $6k$-bit (\textbf{B6}) and hierarchical bracketing (\textbf{HB}, ours) encodings, and the biaffine parser (\textbf{Biaf.}). In the acronyms, subscripts indicate the value of $k$, whereas in the scores, they denote the coverage. The coverage of \textbf{HB} and \textbf{Biaf.} is not indicated because it is always guaranteed to be 100\%. Best SL approach is highlighted in bold, and the best parser is underlined. The English and Czech results correspond to the in-distribution set. The last row ($\mu$) shows the average across all treebanks.}
\end{table*}

\paragraph{Neural tagger} For our experiments, we reproduce the same tagger as in \citet{ezquerro-etal-2024-dependency} to learn the encoding labels and the arc-relation module to learn arc labels from the hidden representations of the predicted arcs. Specifically, we rely on XLM-RoBERTa \cite{conneau-etal-2020-unsupervised} for non-English and XLNet \cite{yang-etal-2019-xlnet} for English treebanks as neural encoders, and two 1-layered FFNs for label and relation prediction. For an input sentence $W=(w_1,..,w_n)\in\mathcal{V}^n$, our encoder computes its contextualized embeddings $(\mathbf{h}_1,...,\mathbf{h}_n)\in\mathbb{R}^{n\times D}$ and uses the first FFN to learn the label distribution for a specific encoding. For inference, the sequence of predicted labels is used to run the decoding process and recover a set of predicted arcs, denoted as $\hat{A}=\{(h\to d): h\in[0,n], d\in[1,n]\}$. To predict the arc relation of each predicted arc, the hidden representations of its head and dependent are concatenated and fed to the second FFN, which learns the arc-relation distribution from these combined representations. To optimize the weights of this second FFN, our framework uses the gold set of arcs at training time to concatenate head and dependent representations. The full architecture is optimized using the cross-entropy losses of both FFNs.

\paragraph{Datasets} We used five datasets from  SemEval 2015 Task 18 \cite{oepen-etal-2015-semeval} with semantic annotations: (i) the English dataset annotated with DELPH-IN MRS-Derived Bi-Lexical Dependencies \citep[DM][]{ivanova-etal-2012-contrastive}, (ii) the English and Chinese datasets with Enju Predicate-Argument Structures \citep[PAS][]{miyao-etal-2005-corpus}, and (iii) the English and Czech datasets with Prague Semantic Dependencies \citep[PSD][]{hajic-etal-2012-announcing}; and the IWPT 2021 Shared Task datasets \cite{bouma-etal-2021-raw} with enhanced dependencies in 17 diverse languages\footnote{We report metrics on the largest treebank of each language, excluding the English and Czech languages since they are already included in the SemEval 2015 Task 18 dataset.}.

\paragraph{Evaluation} We use the SDP evaluation toolkit\footnote{\url{https://github.com/semantic-dependency-parsing/toolkit}.}  \citep{oepen-etal-2015-semeval} to report both the labeled F1 score and exact match (LF, LM). Results with further metrics are included in the Appendix \ref{ap:additional-results}.

\paragraph{Baselines} We run our tagging model with the bracketing and $6k$-bit encodings from \citet{ezquerro-etal-2024-dependency}. For external assessment, we also report the performance of the biaffine parser \cite{dozat-manning-2018-simpler}---a graph-based, non-sequence-labeling model---providing a consistent reference of the state of the art in graph parsing.

\section{Results}
Table \ref{tab:results} shows the performance of different graph encodings and the biaffine baseline. We use acronyms to refer to the bracketing (B) and $6k$-bit (B6) encodings from \citet{ezquerro-etal-2024-dependency} and subindices to specify the value of the hyperparameter $k$. In terms of LF, our hierarchical bracketing encoding (HB) performs competitively, with an average that is only 0.14 below B\textsubscript{3}, 0.26 below B\textsubscript{2}, 0.60 below B6\textsubscript{3}, and 0.71 below the best-performing encoding, B6\textsubscript{4}. Still, a more fine-grained analysis shows that HB outperforms other graph encodings in only two treebanks (English-PAS and Italian-ISDT), although it is still surpassed by the biaffine baseline in both cases. Overall, biaffine obtains the highest LF score (90.45), followed by B6\textsubscript{4} (89.30). 

The outcome shifts when focusing on the exact match performance. Our HB outperforms other approaches, including the biaffine baseline, in 7 treebanks  and obtains the best LM score on average (HB: 38.50), followed by the $6k$-bit encoding (B6\textsubscript{4}: 38.31). When comparing coverage and performance between B6\textsubscript{4} and HB (coverage data are in Table \ref{tab:results} and Tables \ref{tab:en-dm-results}-\ref{tab:uk-iu-results} in the Appendix), we observe that B6\textsubscript{4}'s LM score lags behind in the datasets where its coverage is most limited at $k=4$, like the DM and PAS treebanks, whereas HB handles them better with its guaranteed 100\% coverage.

\begin{figure*}[tpb!]\centering
    \begin{subfigure}[t]{0.29\linewidth}\centering
        \caption{\label{fig:distribution}Rank-frequency distribution.}
        \includegraphics[width=\linewidth]{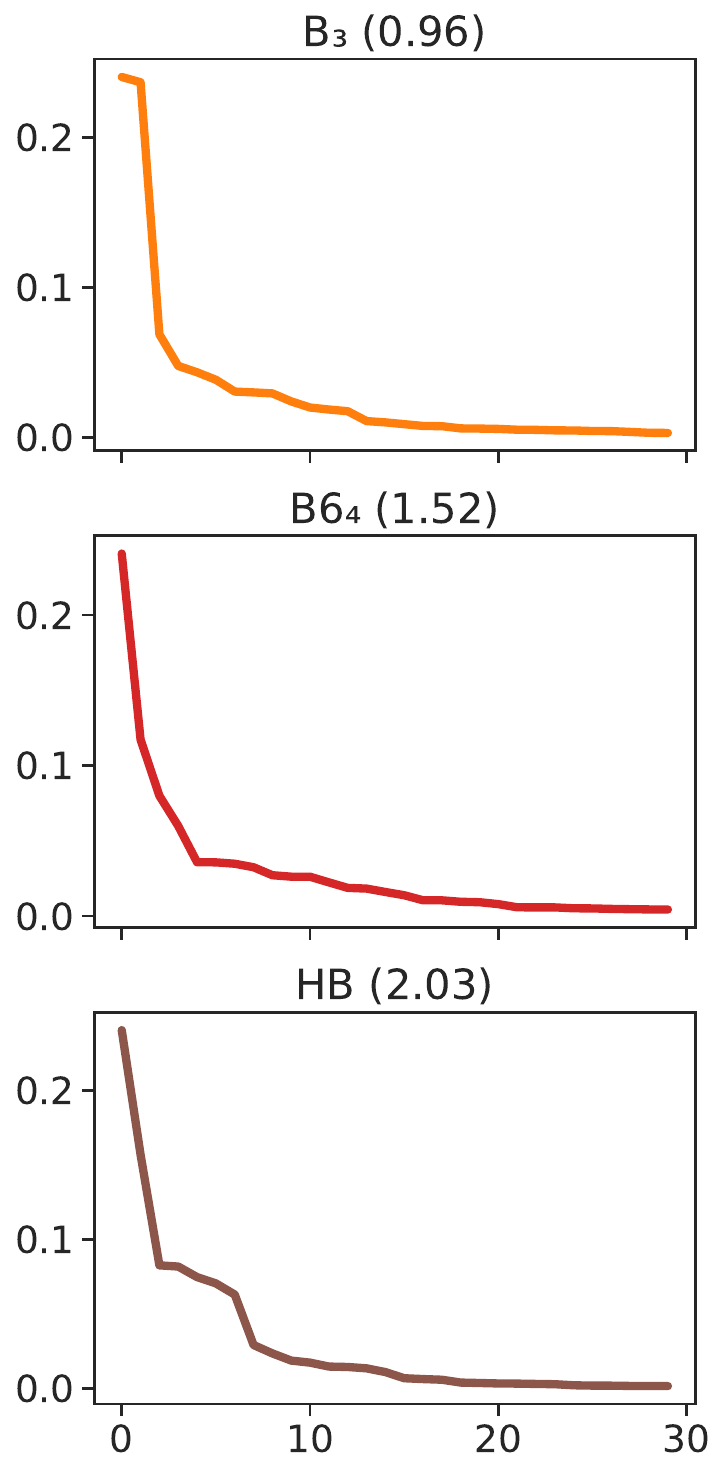}
    \end{subfigure}
    \begin{subfigure}[t]{0.69\linewidth}\centering
        \caption{\label{fig:simple-correlation}Correlation between LM score (y-axis) and $p_{0.5}$ (x-axis).}
        \includegraphics[width=0.89\linewidth]{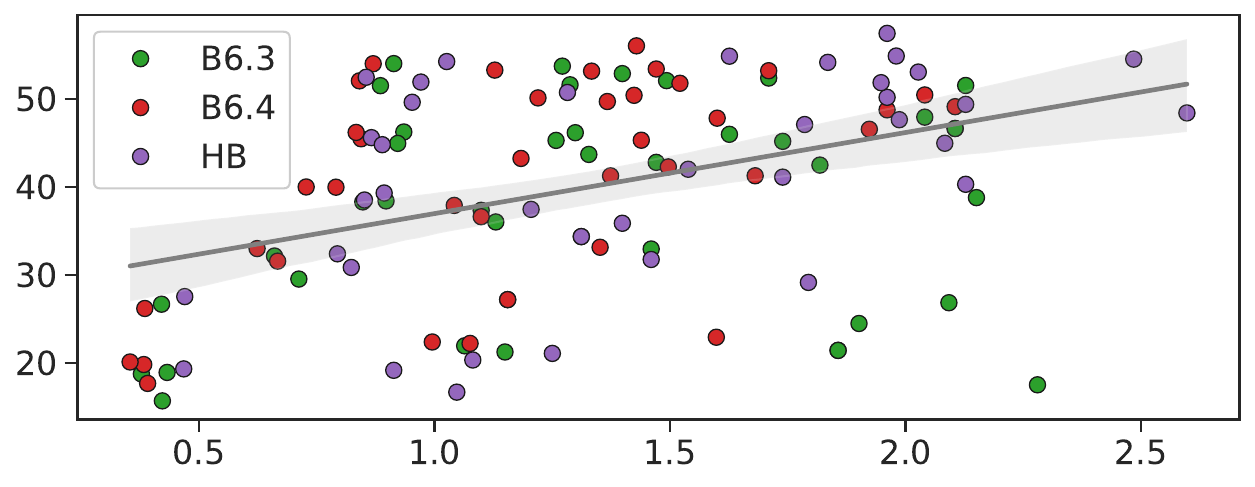}
        \caption{\label{fig:normalized-correlation}Correlation between LM score (y-axis) and normalized $p_{0.5}$ (x-axis).}
        \includegraphics[width=0.89\linewidth]{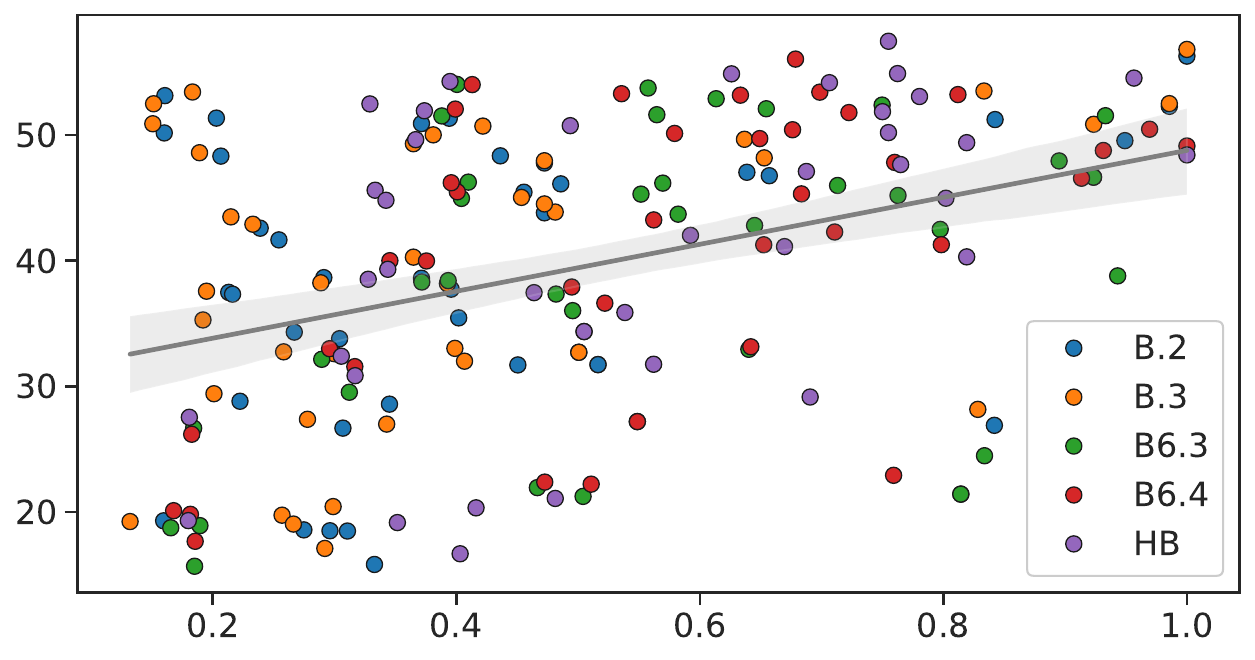}
    \end{subfigure}
    \caption{\label{fig:correlation-analysis}Analysis of the correlation between the encoding performance and the label distribution. Figure \ref{fig:distribution} shows the rank-frequency distribution of labels in the English-DM. Figures \ref{fig:simple-correlation} and \ref{fig:normalized-correlation} visualize the correlation between the LM score and  $p_{0.5}$ across all treebanks.}
\end{figure*}

We also compared the label space of our HB with \citet{ezquerro-etal-2024-dependency}'s approaches. Figure \ref{fig:distribution} shows the rank-frequency distribution of different encoding labels, with a numeric value that indicates the relative rank at the 50th percentile ($p_{0.5}$). For instance,  2.03\% of the most frequent labels of HB account for approximately half of all occurrences. Instead, in B\textsubscript{3}, 0.96\% of the most frequent labels represent half of the occurrences. These values measure the balance of different encodings, in terms of which ratio of the most frequent labels concentrate 50\% of the probability mass. We contrast this value against the LM score in Figure \ref{fig:simple-correlation}, where a significant linear correlation of 88.1\% with a $p$-value $< 0.001$ was detected, indicating a strong dependency between balanced label distributions and exact match performance. We repeated this test adding the scores of the bracketing encodings\footnote{We excluded the bracketing scores from Figure \ref{fig:simple-correlation} since their $p_{0.5}$ follows a different distribution.} but normalizing the values of the $p_{0.5}$ and obtained again a significant correlation of 83.6\% (Figure \ref{fig:normalized-correlation}).

\section{Conclusion}
In this work we apply the non-projective hierarchical bracketing encoding of \citet{ezquerro-etal-2025-hierarchical} to graphs instead of trees. The method offers a competitive trade-off between coverage, label space compactness and parsing performance. Although it does not yield the highest LF score, it achieves the best average exact match across a large multilingual and multi-formalism benchmark. We further showed that the method yields a more balanced label distribution, which correlates strongly with the exact match performance in graph encodings. To our knowledge, this is the first work to empirically demonstrate that hierarchical bracketing encodings can be effectively learned and applied to dependency graph parsing.

\section*{Acknowledgments}

We acknowledge grants GAP (PID2022-139308OA-I00) funded by MICIU/AEI/10.13039/501100011033/ and ERDF, EU; LATCHING (PID2023-147129OB-C21) funded by MICIU/AEI/10.13039/501100011033 and ERDF, EU; and TSI-100925-2023-1 funded by Ministry for Digital Transformation and Civil Service and ``NextGenerationEU'' PRTR; as well as funding by Xunta de Galicia (ED431C 2024/02), and 
CITIC, as a center accredited for excellence within the Galician University System and a member of the CIGUS Network, receives subsidies from the Department of Education, Science, Universities, and Vocational Training of the Xunta de Galicia. Additionally, it is co-financed by the EU through the FEDER Galicia 2021-27 operational program (Ref. ED431G 2023/01).

\section*{Limitations}
\paragraph{Graph formalisms} While a wide range of graph-based annotations have been proposed across NLP tasks -- such as structured sentiment analysis, emotion-cause analysis, and other forms of relational inference -- we focus our experimental study on two well-established formalisms: semantic graph parsing and enhanced dependency parsing. These frameworks offer clear benchmarks and annotation standards, allowing us to evaluate our models in a controlled and widely studied setting.

\paragraph{Computational resources} 
Our experiments were conducted using local high-performance computing infrastructure, with full access to 8 NVIDIA RTX 3090 GPUs (24GB each) and 3 NVIDIA RTX 6000 GPUs (48GB each). These resources allowed us to efficiently train and evaluate all models presented in this work.

\paragraph{Unboundedness} Our proposed encoding is unbounded, meaning that the number of labels needed to encode a given dataset or tree is not theoretically bounded by a constant. This is a purely theoretical limitation, but our experiments show that it is not relevant in practice for the dataset tested, as in practice it requires fewer labels than the bounded encodings B6\textsubscript{3} and B6\textsubscript{4} (Table~\ref{tab:stats-labels} in the Appendix).

\section*{Ethical considerations} Our work benchmarks the hierarchical bracketing encoding proposed by \citet{ezquerro-etal-2025-hierarchical} in the context of dependency graph parsing. None of the materials used involve sensitive personal data, human subjects, or contexts that might pose ethical risks. Consequently, the techniques can be incorporated into research without ethical reservations.

We acknowledge the environmental impact of training neural models, particularly with respect to CO\textsubscript{2} emissions. All experiments were conducted in Spain, where we measured the carbon footprint during both training and inference. Training produces around 0.28 g CO\textsubscript{2} per epoch and inference approximately 0.19 g CO\textsubscript{2}. These values remain low compared to recent large NLP models. For comparison, the European Union sets a limit of roughly 115 g CO\textsubscript{2} per kilometer for newly manufactured cars.

\bibliography{custom}
\appendix

\section{Appendix}\label{sec:appendix}
\subsection{Treebank statistics}
This work conducts experiments in all treebanks of the SemEval 2015 Task 18 dataset \cite{oepen-etal-2015-semeval} and a selection of IWPT datasets \cite{bouma-etal-2021-raw}. Tables \ref{tab:stats-sdp} and \ref{tab:stats-iwpt} summarize different statistics of each treebank. Table \ref{tab:stats-labels} shows information about the labels generated in each encoding.
\begin{table}[h]\footnotesize\centering 
    \setlength{\tabcolsep}{2pt}
    \renewcommand{\arraystretch}{1}
    \begin{tabular}{|c|cc|ccc|ccc|}
        \cline{2-9}
        \multicolumn{1}{c|}{}& \multirow{2}{*}{\textbf{\#sents}} & \multirow{2}{*}{$n$}  & \multicolumn{3}{c|}{\textbf{\%r.planes}} & \multirow{2}{*}{\textbf{d.}} & \multirow{2}{*}{\textbf{$|R|$}} & \multirow{2}{*}{\textbf{\#cycs}}\\
        \multicolumn{1}{c|}{}&&& 1 & 2 & 3 & &&\\
        \hline 
        \textit{en}\textsubscript{DM} & 33964 & 22.52 & 86.28 & 13.68 & 0.05 & 0.78 & 9.87 & 0 \\
        & 1692 & 22.28 & 86.94 & 13.06 & 0.00 & 0.79 & 9.79 & 0 \\
        & 1410 & 22.66 & 84.11 & 15.89 & 0.00 & 0.77 & 9.80 & 0 \\
        & 1849 & 17.08 & 88.64 & 11.20 & 0.16 & 0.75 & 7.26 & 0 \\
        \hline 
        \textit{en}\textsubscript{PAS} & 33964 & 22.52 & 83.70 & 16.20 & 0.10 & 1.01 & 10.95 & 0 \\
        & 1692 & 22.28 & 86.52 & 13.42 & 0.06 & 1.00 & 10.86 & 0 \\
        & 1410 & 22.66 & 82.91 & 17.02 & 0.07 & 1.01 & 10.92 & 0 \\
        & 1849 & 17.08 & 83.02 & 16.77 & 0.22 & 0.99 & 8.35 & 0 \\
        \hline
        \textit{en}\textsubscript{PSD} & 33964 & 22.52 & 77.51 & 20.99 & 1.44 & 0.70 & 7.89 & 0 \\
        & 1692 & 22.28 & 76.71 & 22.16 & 0.95 & 0.70 & 7.84 & 0 \\
        & 1410 & 22.66 & 77.38 & 21.21 & 1.28 & 0.69 & 7.81 & 0 \\
        & 1849 & 17.08 & 81.23 & 17.79 & 0.87 & 0.67 & 5.59 & 0 \\
        \hline
        \textit{cs}\textsubscript{PSD} & 40047 & 23.45 & 74.22 & 24.04 & 1.66 & 0.77 & 8.90 & 0 \\
        & 2010 & 22.99 & 75.32 & 23.33 & 1.24 & 0.78 & 8.82 & 0 \\
        & 1670 & 22.99 & 73.35 & 24.67 & 1.98 & 0.76 & 8.53 & 0 \\
        & 5226 & 16.82 & 78.66 & 19.15 & 2.01 & 0.78 & 6.31 & 0 \\
        \hline 
        \textit{zh}\textsubscript{PAS} & 25896 & 22.43 & 75.60 & 24.12 & 0.28 & 1.02 & 11.48 & 0 \\
         & 2440 & 27.95 & 73.16 & 26.60 & 0.25 & 1.02 & 14.63 & 0 \\
         & 8976 & 23.89 & 75.12 & 24.64 & 0.23 & 1.02 & 12.26 & 0 \\
        \hline 
    \end{tabular}
    \caption{\label{tab:stats-sdp}Treebank statistics for the SDP datasets. Number of sentences (\textbf{\#sents}), average sentence length ($n$), distribution of sentences by number of relaxed planes (\textbf{\%r.planes}), density (\textbf{d}) as the average ratio of arcs and nodes per graph, average number of structural arcs (\textbf{$|R|$}) and number of cycles \textbf{(\#cycs.}). Different splits in each subrow: train, development, ID and OOD or test.}
\end{table}

\begin{table}[h]\centering\footnotesize
    \setlength{\tabcolsep}{2pt}
    \renewcommand{\arraystretch}{1.1}
    \begin{tabular}{|c|cc|ccc|ccc|}
        \cline{2-9}
        \multicolumn{1}{c|}{}& \multirow{2}{*}{\textbf{\#sents}} & \multirow{2}{*}{$n$}  & \multicolumn{3}{c|}{\textbf{\%r.planes}} & \multirow{2}{*}{\textbf{d.}} & \multirow{2}{*}{\textbf{$|R|$}} & \multirow{2}{*}{\textbf{\#cycs}}\\
        \multicolumn{1}{c|}{}&&& 1 & 2 & 3 & &&\\
        \hline 
        \textit{ar}\textsubscript{PADT} &  6075 & 36.85 & 65.32 & 32.49 & 1.86 & 1.53 & 18.95 & 1783 \\
        & 909 & 33.27 & 69.97 & 28.49 & 1.32 & 1.05 & 17.05 & 282 \\
        & 680 & 41.56 & 64.26 & 33.82 & 1.76 & 1.05 & 21.31 & 235 \\
        \hline
        \textit{bg}\textsubscript{BTB}  & 8907 & 13.96 & 90.14 & 9.68 & 0.18 & 1.02 & 6.41 & 1177 \\
        & 1115 & 14.43 & 89.69 & 10.22 & 0.09 & 1.02 & 6.63 & 153 \\
        & 1116 & 14.09 & 90.59 & 9.41 & 0.00 & 1.02 & 6.44 & 145 \\
        \hline 
        \textit{fi}\textsubscript{TDT}  & 12217 & 13.33 & 78.90 & 19.19 & 1.77 & 1.06 & 6.29 & 2906 \\
        & 1364 & 13.42 & 78.30 & 20.01 & 1.32 & 1.06 & 6.31 & 340 \\
        &  1555 & 13.55 & 77.17 & 20.77 & 1.93 & 1.07 & 6.34 & 357 \\
        \hline 
        \textit{fr}\textsubscript{SEQ.} & 2231 & 22.64 & 80.55 & 19.23 & 0.18 & 1.04 & 9.45 & 770 \\
        & 412 & 24.28 & 80.83 & 18.93 & 0.24 & 1.04 & 10.00 & 166 \\
        & 456 & 22.04 & 79.82 & 19.74 & 0.44 & 1.04 & 9.05 & 119 \\
        \hline 
        \textit{it}\textsubscript{ISDT}  & 13121 & 21.04 & 85.73 & 14.17 & 0.10 & 1.03 & 9.19 & 2963 \\
        & 564 & 21.11 & 87.06 & 12.94 & 0.00 & 1.03 & 9.27 & 136 \\
         & 482 & 21.61 & 86.31 & 13.49 & 0.21 & 1.03 & 9.40 & 109 \\
         \hline 
        \textit{lt}\textsubscript{ALK.}  & 2341 & 20.35 & 51.82 & 44.55 & 3.03 & 1.10 & 9.77 & 388 \\
        & 617 & 18.74 & 58.67 & 39.87 & 1.46 & 1.07 & 9.11 & 89 \\
        & 684 & 15.86 & 53.22 & 44.01 & 2.34 & 1.08 & 7.42 & 63 \\
        \hline 
        \textit{lv}\textsubscript{LVTB} &  10156 & 16.50 & 73.50 & 24.25 & 2.10 & 1.06 & 7.60 & 2189 \\
        & 1664 & 15.60 & 75.12 & 22.84 & 1.86 & 1.04 & 7.13 & 255 \\
        & 1823 & 14.48 & 78.72 & 19.69 & 1.37 & 1.02 & 6.66 & 297 \\
        \hline 
        \textit{nl}\textsubscript{ALP.}  & 12264 & 15.16 & 82.89 & 16.47 & 0.63 & 1.03 & 6.37 & 1575 \\
        & 718 & 16.07 & 87.47 & 11.84 & 0.70 & 1.02 & 6.78 & 62 \\
        & 596 & 18.53 & 76.17 & 22.99 & 0.84 & 1.04 & 7.68 & 68 \\
        \hline 
        \textit{pl}\textsubscript{PDB} & 17722 & 15.90 & 67.57 & 30.98 & 1.37 & 1.06 & 7.55 & 2677 \\
        & 2215 & 15.66 & 68.04 & 30.61 & 1.35 & 1.06 & 7.46 & 322 \\
        &  2215 & 15.18 & 68.53 & 30.16 & 1.13 & 1.06 & 7.27 & 351 \\
        \hline 
        \textit{ru}\textsubscript{SYN.} & 48814 & 17.83 & 67.29 & 32.30 & 0.41 & 1.04 & 8.41 & 4560 \\
        & 6584 & 18.00 & 65.31 & 34.23 & 0.44 & 1.05 & 8.37 & 580 \\
        & 6491 & 18.08 & 65.23 & 34.42 & 0.35 & 1.05 & 8.52 & 588 \\
        \hline 
        \textit{sk}\textsubscript{SNK} & 8483 & 9.50 & 78.52 & 21.10 & 0.38 & 1.04 & 4.37 & 565 \\
        & 1060 & 12.01 & 81.13 & 18.40 & 0.47 & 1.05 & 5.80 & 120 \\
        & 1061 & 12.00 & 77.38 & 22.34 & 0.28 & 1.05 & 5.78 & 164 \\
        \hline 
        \textit{sv}\textsubscript{TAL.} & 4303 & 15.49 & 85.99 & 13.97 & 0.05 & 1.05 & 6.52 & 812 \\
        & 504 & 19.44 & 76.19 & 23.61 & 0.20 & 1.06 & 7.80 & 138 \\
        & 1219 & 16.72 & 85.23 & 14.60 & 0.16 & 1.05 & 6.95 & 242 \\
        \hline 
        \textit{ta}\textsubscript{TBT} &  400 & 15.82 & 97.75 & 2.25 & 0.00 & 1.02 & 8.21 & 1 \\
        & 80 & 15.79 & 98.75 & 1.25 & 0.00 & 1.05 & 8.29 & 25 \\
        & 120 & 16.57 & 98.33 & 1.67 & 0.00 & 1.03 & 8.41 & 43 \\
        \hline
        \textit{uk}\textsubscript{IU} & 5496 & 16.81 & 63.36 & 35.94 & 0.69 & 1.06 & 7.71 & 968 \\
        & 672 & 18.71 & 61.61 & 37.80 & 0.60 & 1.07 & 8.83 & 198 \\
        & 892 & 19.19 & 65.25 & 33.86 & 0.90 & 1.05 & 9.07 & 152 \\
        \hline 
    \end{tabular}
    \caption{\label{tab:stats-iwpt} Treebank statistics for the IWPT datasets. The notation is the same as in Table \ref{tab:stats-sdp}.}
\end{table}

\begin{table}[h]\centering\footnotesize
    \setlength{\tabcolsep}{2pt}
    \renewcommand{\arraystretch}{1.1}
    \begin{tabular}{|c|cc|cc|cc|cc|cc|}
        \cline{2-11}
        \multicolumn{1}{c|}{}& \multicolumn{2}{c|}{\textbf{B\textsubscript{2}}} & \multicolumn{2}{c|}{\textbf{B\textsubscript{3}}} & \multicolumn{2}{c|}{\textbf{B6\textsubscript{3}}} & \multicolumn{2}{c|}{\textbf{B6\textsubscript{4}}} & \multicolumn{2}{c|}{\textbf{HB}}\\
        \hline 
        \textit{en}\textsubscript{DM}& 466 & 21 & 483 & 23 & 486 & 25 & 621 & 29 & 319 & 8 \\
        \textit{en}\textsubscript{PAS} & 900 & 42 & 929 & 45 & 543 & 19 & 905 & 44 & 356 & 12 \\
        \textit{en}\textsubscript{PSD} & 1013 & 51 & 1265 & 87 & 681 & 39 & 823 & 51 & 650 & 57 \\
        \textit{cs}\textsubscript{PSD} & 1638 & 140 & 2021 & 204 & 856 & 67 & 1047 & 92 & 918 & 80 \\
        \textit{zh}\textsubscript{PAS} & 1070 & 184 & 1117 & 198 & 474 & 36 & 774 & 85 & 377 & 49 \\
        \hline 
        \textit{ar}\textsubscript{PADT} & 830 & 81 & 971 & 105 & 368 & 23 & 430 & 36 & 369 & 36 \\
        \textit{bg}\textsubscript{BTB} & 457 & 43 & 478 & 44 & 199 & 18 & 220 & 19 & 226 & 17 \\
        \textit{cs}\textsubscript{PDT} & 2128 & 213 & 2667 & 300 & 801 & 54 & 1015 & 84 & 958 & 115 \\
        \textit{en}\textsubscript{EWT} & 867 & 52 & 902 & 59 & 263 & 11 & 297 & 16 & 293 & 11 \\
        \textit{et}\textsubscript{EDT} & 391 & 37 & 391 & 37 & 118 & 12 & 119 & 12 & 114 & 13 \\
        \textit{fi}\textsubscript{TDT} & 1019 & 90 & 1259 & 122 & 415 & 21 & 507 & 32 & 532 & 60 \\
        \textit{fr}\textsubscript{SEQ.} & 486 & 78 & 497 & 81 & 168 & 19 & 188 & 20 & 163 & 20 \\
        \textit{it}\textsubscript{ISDT} & 859 & 18 & 881 & 19 & 254 & 5 & 280 & 5 & 242 & 9 \\
        \textit{lt}\textsubscript{ALK.} & 769 & 109 & 923 & 142 & 352 & 39 & 411 & 50 & 388 & 58 \\
        \textit{lv}\textsubscript{LVTB} & 994 & 124 & 1236 & 176 & 396 & 43 & 486 & 60 & 515 & 55 \\
        \textit{nl}\textsubscript{ALP.} & 758 & 34 & 845 & 39 & 291 & 8 & 311 & 10 & 306 & 3 \\
        \textit{pl}\textsubscript{PDB} & 1130 & 112 & 1381 & 160 & 458 & 35 & 549 & 49 & 515 & 48 \\
        \textit{ru}\textsubscript{SYN.} & 1212 & 108 & 1358 & 131 & 350 & 23 & 383 & 26 & 345 & 23 \\
        \textit{sk}\textsubscript{SNK} & 452 & 74 & 490 & 85 & 220 & 32 & 233 & 40 & 199 & 32 \\
        \textit{sv}\textsubscript{TAL.} & 560 & 110 & 563 & 115 & 203 & 25 & 225 & 32 & 204 & 27 \\
        \textit{ta}\textsubscript{TTB} & 92 & 26 & 92 & 26 & 33 & 12 & 33 & 12 & 28 & 8 \\
        \textit{uk}\textsubscript{IU} & 787 & 102 & 849 & 118 & 337 & 30 & 374 & 33 & 323 & 28 \\
         \hline 
         $\mu$ & 858 & 84 & 982 & 105 & 376 & 27 & 465 & 38 & 379 & 35 \\ 
         \hline 
    \end{tabular}
    \caption{\label{tab:stats-labels}Number of generated labels in the training set and number of unseen labels in the development and evaluation sets. Average in the last row ($\mu$).}
\end{table}

\subsection{Training configuration}\label{ap:training-configuration}
All our models were trained using the AdamW optimizer \cite{loshchilov-hutter-2019-decoupled} with a learning rate of $\eta=10^{-5}$, for 100 epochs and with token-based batching set to 500 tokens. The FFN layers use the LeakyReLU activation \cite{xu-etal-2015-empirical} with a negative slope of 0.1 and a dropout rate set to 0.1. The evaluation on the development set was used as the stopping criterion.

\subsection{Additional results}\label{ap:additional-results}
Tables \ref{tab:en-dm-results} to \ref{tab:uk-iu-results} show the detailed performance of each encoding in the evaluation sets. We use a similar notation to Table \ref{tab:results}, including the (un)labeled F1-score (\textbf{*F}) and exact match (\textbf{*M}), and the tag accuracy (\textbf{A}) and ratio of well-formed graphs (\textbf{W}). 

\begin{table*}[h!]\centering\footnotesize    
    \setlength{\tabcolsep}{2pt}
    \renewcommand{\arraystretch}{1.1}
    \begin{tabular}{|c|cccc|cc|cccc|cc|cccc|cc|}
        \cline{2-19}
        \multicolumn{1}{c|}{}& \multicolumn{6}{c|}{\textbf{dev}} & \multicolumn{6}{c|}{\textbf{id}} & \multicolumn{6}{c|}{\textbf{ood}}\\
        \cline{2-19}
        \multicolumn{1}{c|}{}& UF & LF & UM & LM & A & W & UF & LF & UM & LM & A & W & UF & LF & UM & LM & A & W\\ 
        \hline 
        \textbf{B\textsubscript{2}} & 100 & 100 & 100 & 100 & - & - & 100 & 100 & 100 & 100 & - & - & *100 & *100 & 99.84 & 99.84 & - & - \\
        \textbf{B\textsubscript{3}} & 100 & 100 & 100 & 100 & - & - & 100 & 100 & 100 & 100 & - & - & 100 & 100 & 100 & 100 & - & - \\
        \textbf{B6\textsubscript{3}} & 99.38 & 99.38 & 79.85 & 79.85 & - & - & 99.51 & 99.51 & 83.62 & 83.62 & - & - & 99.66 & 99.66 & 89.89 & 89.89 & - & - \\
        \textbf{B6\textsubscript{4}} & 99.95 & 99.95 & 97.93 & 97.93 & - & - & 99.96 & 99.96 & 98.51 & 98.51 & - & - & 99.97 & 99.97 & 98.97 & 98.97 & - & - \\
        \textbf{HB} & 100 & 100 & 100 & 100 & - & - & 100 & 100 & 100 & 100 & - & - & 100 & 100 & 100 & 100 & - & - \\
        \hline  
        \textbf{B\textsubscript{2}} & 95.56 & 95.10 & 59.04 & 56.26 & 94.53 & 85.40 & 95.30 & 94.57 & 56.52 & 52.27 & \textbf{\underline{94.55}} & 82.12 & 92.58 & 92.15 & 52.79 & 49.54 & 91.43 & 75.93 \\
        \textbf{B\textsubscript{3}} & \textbf{95.68} & \textbf{95.25} & 60.11 & 56.80 & \textbf{\underline{94.73}} & 85.51 & 95.28 & 94.67 & 56.45 & 52.48 & 94.53 & 84.29 & \textbf{92.75} & 92.19 & 53.76 & \textbf{\underline{50.84}} & \textbf{\underline{91.59}} & 79.82 \\
        \textbf{B6\textsubscript{3}} & 94.66 & 94.25 & 47.87 & 46.16 & 93.42 & 79.38 & 94.59 & 94.04 & 46.88 & 43.69 & 93.32 & 78.90 & 91.84 & 91.58 & 48.30 & 45.97 & 90.08 & 71.44 \\
        \textbf{B6\textsubscript{4}} & 95.62 & 95.13 & 59.16 & 56.03 & 93.66 & 82.32 & \textbf{95.37} & \textbf{94.74} & 55.39 & 51.77 & 93.44 & 76.86 & 92.67 & \textbf{92.28} & 53.76 & 50.41 & 90.45 & 73.25 \\
        \textbf{HB} & 95.60 & 95.10 & \textbf{\underline{61.35}} & \textbf{\underline{57.45}} & 93.99 & \textbf{\underline{90.01}} & 95.07 & 94.52 & \textbf{\underline{57.16}} & \textbf{\underline{53.05}} & 93.66 & \textbf{\underline{89.66}} & 92.39 & 92.09 & \textbf{\underline{54.14}} & 50.73 & 90.55 & \textbf{\underline{85.55}} \\
        \hline 
        \textbf{Biaf} & \underline{95.93} & \underline{95.76} & 54.79 & 51.83 & - & - & \underline{95.51} & \underline{95.45} & 50.00 & 46.95 & - & - & \underline{92.82} & \underline{92.92} & 48.08 & 45.65 & - & - \\
         \hline 
    \end{tabular}
    \caption{\label{tab:en-dm-results}Performance on the English-DM (\textit{en}\textsubscript{DM}) dataset. The first and second row groups correspond to coverage and parsing metrics, respectively. An asterisk (*100) denotes coverage metrics that are near full coverage. Acronyms follow the notation used in Table \ref{tab:results}.}
\end{table*}

\begin{table*}[h]\centering\footnotesize    
    \setlength{\tabcolsep}{2pt}
    \renewcommand{\arraystretch}{1.1}
    \begin{tabular}{|c|cccc|cc|cccc|cc|cccc|cc|}
        \cline{2-19}
        \multicolumn{1}{c|}{}& \multicolumn{6}{c|}{\textbf{dev}} & \multicolumn{6}{c|}{\textbf{id}} & \multicolumn{6}{c|}{\textbf{ood}}\\
        \cline{2-19}
        \multicolumn{1}{c|}{}& UF & LF & UM & LM & A & W & UF & LF & UM & LM & A & W & UF & LF & UM & LM & A & W\\ 
        \hline 
        \textbf{B\textsubscript{2}} & *100 & *100 & 99.94 & 99.94 & - & - & *100 & *100 & 99.93 & 99.93 & - & - & *100 & *100 & 99.78 & 99.78 & - & - \\
        \textbf{B\textsubscript{3}} & 100 & 100 & 100 & 100 & - & - & 100 & 100 & 100 & 100 & - & - & 100 & 100 & 100 & 100 & - & - \\
        \textbf{B6\textsubscript{3}} & 97.73 & 97.73 & 46.45 & 46.45 & - & - & 97.56 & 97.56 & 42.55 & 42.55 & - & - & 98.24 & 98.24 & 63.33 & 63.33 & - & - \\
        \textbf{B6\textsubscript{4}} & 99.39 & 99.39 & 80.08 & 80.08 & - & - & 99.37 & 99.37 & 78.30 & 78.30 & - & - & 99.53 & 99.53 & 87.29 & 87.29 & - & - \\
        \textbf{HB} & 100 & 100 & 100 & 100 & - & - & 100 & 100 & 100 & 100 & - & - & 100 & 100 & 100 & 100 & - & - \\
        \hline 
        \textbf{B\textsubscript{2}} & 96.38 & 95.40 & 55.26 & 46.75 & 93.54 & 85.37 & 96.08 & 95.27 & 54.33 & 47.02 & 93.52 & 83.00 & 94.76 & 93.80 & 56.90 & 51.22 & 91.96 & 81.11 \\
        \textbf{B\textsubscript{3}} & \textbf{96.50} & \textbf{95.63} & 56.62 & \textbf{\underline{48.17}} & \textbf{\underline{93.63}} & 89.96 & \textbf{96.32} & 95.50 & 57.30 & 49.65 & \textbf{\underline{93.76}} & 89.27 & \textbf{95.11} & \textbf{94.13} & 58.79 & 53.49 & \textbf{\underline{92.16}} & 82.58 \\
        \textbf{B6\textsubscript{3}} & 94.25 & 93.80 & 30.50 & 26.83 & 92.52 & 86.03 & 93.90 & 93.56 & 27.59 & 24.47 & 92.27 & 84.34 & 93.41 & 92.54 & 42.02 & 38.78 & 90.93 & 79.07 \\
        \textbf{B6\textsubscript{4}} & 95.95 & 95.13 & 48.70 & 41.25 & 92.34 & 90.15 & 95.66 & 95.01 & 48.01 & 42.27 & 91.81 & 87.70 & 94.45 & 93.59 & 53.70 & 49.70 & 90.22 & 81.29 \\
        \textbf{HB} & 96.32 & 95.32 & \textbf{\underline{57.68}} & 47.64 & 92.82 & \textbf{\underline{96.61}} & 96.26 & \textbf{95.59} & \textbf{\underline{59.08}} & \textbf{\underline{51.84}} & 93.31 & \textbf{\underline{95.48}} & 94.96 & 94.01 & \textbf{\underline{60.09}} & \textbf{54.52} & 91.54 & \textbf{\underline{92.53}} \\
        \hline 
        \textbf{Biaf} & \underline{96.69} & \underline{95.87} & 55.73 & 47.10 & - & - & \underline{96.64} & \underline{96.13} & 55.96 & 48.51 & - & - & \underline{95.75} & \underline{94.97} & 60.03 & \underline{54.73} & - & - \\
        \hline 
    \end{tabular}
    \caption{\label{tab:en-pas-results}Performance in the English-PAS (\textit{en}\textsubscript{PAS}) dataset. Same notation as in Table \ref{tab:en-dm-results}.}
\end{table*}

\begin{table*}[h]\centering\footnotesize    
    \setlength{\tabcolsep}{2pt}
    \renewcommand{\arraystretch}{1.1}
    \begin{tabular}{|c|cccc|cc|cccc|cc|cccc|cc|}
        \cline{2-19}
        \multicolumn{1}{c|}{}& \multicolumn{6}{c|}{\textbf{dev}} & \multicolumn{6}{c|}{\textbf{id}} & \multicolumn{6}{c|}{\textbf{ood}}\\
        \cline{2-19}
        \multicolumn{1}{c|}{}& UF & LF & UM & LM & A & W & UF & LF & UM & LM & A & W & UF & LF & UM & LM & A & W\\ 
        \hline 
        \textbf{B\textsubscript{2}} & 99.97 & 99.97 & 98.88 & 98.88 & - & - & 99.96 & 99.96 & 98.58 & 98.58 & - & - & 99.97 & 99.97 & 99.03 & 99.03 & - & - \\
        \textbf{B\textsubscript{3}} & *100 & *100 & 99.82 & 99.82 & - & - & *100 & *100 & 99.86 & 99.86 & - & - & *100 & *100 & 99.89 & 99.89 & - & - \\
        \textbf{B6\textsubscript{3}} & 99.97 & 99.97 & 98.94 & 98.94 & - & - & 99.98 & 99.98 & 98.94 & 98.94 & - & - & 99.96 & 99.96 & 98.81 & 98.81 & - & - \\
        \textbf{B6\textsubscript{4}} & 99.99 & 99.99 & 99.65 & 99.65 & - & - & *100 & *100 & 99.93 & 99.93 & - & - & 99.99 & 99.99 & 99.68 & 99.68 & - & - \\
        \textbf{HB} & 100 & 100 & 100 & 100 & - & - & 100 & 100 & 100 & 100 & - & - & 100 & 100 & 100 & 100 & - & - \\
        \hline 
        \textbf{B\textsubscript{2}} & 93.74 & 86.93 & 48.76 & 18.50 & 92.44 & 75.64 & 92.53 & 85.33 & 45.60 & 15.82 & 91.29 & 73.52 & 92.42 & 86.09 & 52.46 & 26.66 & 90.47 & 72.79 \\
        \textbf{B\textsubscript{3}} & 94.17 & 87.21 & 51.89 & 19.03 & 92.73 & 81.97 & 93.10 & 85.95 & 47.09 & 17.09 & 91.52 & 79.77 & 92.81 & \textbf{86.60} & 54.68 & 27.37 & 90.81 & 79.09 \\
        \textbf{B6\textsubscript{3}} & 94.15 & 87.14 & 51.18 & 18.74 & 93.45 & 84.10 & 93.06 & 85.68 & 46.81 & 15.67 & 92.52 & 77.96 & 92.83 & 86.52 & 53.81 & 26.66 & 91.91 & 79.89 \\
        \textbf{B6\textsubscript{4}} & \textbf{94.37} & \textbf{87.54} & \textbf{\underline{53.19}} & \textbf{\underline{20.09}} & \textbf{\underline{93.61}} & 86.89 & \textbf{93.37} & \textbf{86.20} & \textbf{\underline{48.44}} & \textbf{\underline{17.66}} & \textbf{\underline{92.80}} & 82.55 & \textbf{93.08} & 86.56 & \textbf{\underline{55.11}} & 26.18 & \textbf{\underline{92.01}} & 83.84 \\
        \textbf{HB} & 93.64 & 86.85 & 52.01 & 19.15 & 92.87 & \textbf{\underline{91.21}} & 92.52 & 85.61 & 47.73 & 16.67 & 92.22 & \textbf{\underline{88.66}} & 91.91 & 85.90 & 53.43 & \textbf{\underline{27.53}} & 91.33 & \textbf{\underline{89.02}} \\
        \hline 
        \textbf{Biaf} & \underline{94.64} & \underline{87.91} & 47.75 & 19.56 & - & - & \underline{93.70} & \underline{86.84} & 44.47 & 16.60 & - & - & \underline{93.12} & \underline{87.00} & 50.84 & 26.55 & - & - \\
        \hline 
    \end{tabular}
    \caption{\label{tab:en-psd-results}Performance in the English-PSD (\textit{en}\textsubscript{PSD}) dataset. Same notation as in Table \ref{tab:en-dm-results}.}
\end{table*}

\begin{table*}[h]\centering\footnotesize    
    \setlength{\tabcolsep}{2pt}
    \renewcommand{\arraystretch}{1.1}
    \begin{tabular}{|c|cccc|cc|cccc|cc|cccc|cc|}
        \cline{2-19}
        \multicolumn{1}{c|}{}& \multicolumn{6}{c|}{\textbf{dev}} & \multicolumn{6}{c|}{\textbf{id}} & \multicolumn{6}{c|}{\textbf{ood}}\\
        \cline{2-19}
        \multicolumn{1}{c|}{}& UF & LF & UM & LM & A & W & UF & LF & UM & LM & A & W & UF & LF & UM & LM & A & W\\ 
        \hline 
        \textbf{B\textsubscript{2}} & 99.96 & 99.96 & 98.66 & 98.66 & - & - & 99.96 & 99.96 & 98.02 & 98.02 & - & - & 99.94 & 99.94 & 97.82 & 97.82 & - & - \\
        \textbf{B\textsubscript{3}} & *100 & *100 & 99.90 & 99.90 & - & - & 100 & 100 & 100 & 100 & - & - & *100 & *100 & 99.83 & 99.83 & - & - \\
        \textbf{B6\textsubscript{3}} & 99.98 & 99.98 & 99.35 & 99.35 & - & - & 99.98 & 99.98 & 99.22 & 99.22 & - & - & 99.97 & 99.97 & 98.91 & 98.91 & - & - \\
        \textbf{B6\textsubscript{4}} & 99.99 & 99.99 & 99.85 & 99.85 & - & - & *100 & *100 & 99.82 & 99.82 & - & - & 99.99 & 99.99 & 99.69 & 99.69 & - & - \\
        \textbf{HB} & 100 & 100 & 100 & 100 & - & - & 100 & 100 & 100 & 100 & - & - & 100 & 100 & 100 & 100 & - & - \\
        \hline 
        \textbf{B\textsubscript{2}} & 94.49 & 90.84 & 48.16 & 31.69 & 91.87 & 76.10 & 93.75 & 90.02 & 45.81 & 28.80 & 91.34 & 72.98 & 91.85 & 83.12 & 49.00 & 19.29 & 87.95 & 71.32 \\
        \textbf{B\textsubscript{3}} & 94.44 & 90.73 & 47.91 & 31.99 & 91.85 & 76.87 & 93.60 & 90.03 & 45.99 & 29.40 & 91.08 & 75.31 & 91.82 & 83.07 & 48.99 & 19.23 & 87.73 & 71.23 \\
        \textbf{B6\textsubscript{3}} & 94.89 & \textbf{91.29} & 49.85 & 32.14 & \textbf{\underline{93.05}} & 76.65 & 94.01 & 90.25 & 46.53 & 29.52 & 92.15 & 76.57 & 91.67 & 83.00 & 49.06 & 18.91 & 89.05 & 73.99 \\
        \textbf{B6\textsubscript{4}} & \textbf{94.94} & 91.15 & \textbf{\underline{51.14}} & \textbf{\underline{32.99}} & 93.00 & 82.45 & \textbf{94.15} & \textbf{90.48} & \textbf{\underline{48.32}} & \textbf{\underline{31.56}} & \textbf{\underline{92.31}} & 80.56 & \textbf{92.28} & \textbf{83.54} & \textbf{\underline{51.19}} & \textbf{\underline{19.80}} & \textbf{\underline{89.23}} & 78.53 \\
        \textbf{HB} & 93.84 & 90.20 & 49.00 & 32.39 & 91.95 & \textbf{\underline{86.97}} & 93.29 & 89.64 & 47.90 & 30.84 & 91.36 & \textbf{\underline{87.28}} & 91.01 & 82.35 & 49.16 & 19.31 & 88.90 & \textbf{\underline{85.60}} \\
        \hline
        \textbf{Biaf} & \underline{95.30} & \underline{92.07} & 44.03 & 30.20 & - & - & \underline{94.61} & \underline{91.21} & 43.11 & 27.90 & - & - & \underline{92.55} & \underline{83.81} & 44.18 & 17.76 & - & - \\
        \hline 
    \end{tabular}
    \caption{\label{tab:cs-psd-results}Performance in the Czech-PSD (\textit{cs}\textsubscript{PSD}) dataset. Same notation as in Table \ref{tab:en-dm-results}.}
\end{table*}

\begin{table*}[h]\centering\footnotesize    
    \setlength{\tabcolsep}{2pt}
    \renewcommand{\arraystretch}{1.1}
    \begin{tabular}{|c|cccc|cc|cccc|cc|}
        \cline{2-13}
        \multicolumn{1}{c|}{}& \multicolumn{6}{c|}{\textbf{dev}} & \multicolumn{6}{c|}{\textbf{id}} \\
        \cline{2-13}
        \multicolumn{1}{c|}{}& UF & LF & UM & LM & A & W & UF & LF & UM & LM & A & W \\
        \hline 
        \textbf{B\textsubscript{2}} & *100 & *100 & 99.75 & 99.75 & - & - & *100 & *100 & 99.77 & 99.77 & - & - \\
        \textbf{B\textsubscript{3}} & 100 & 100 & 100 & 100 & - & - & 100 & 100 & 100 & 100 & - & - \\
        \textbf{B6\textsubscript{3}} & 96.65 & 96.65 & 43.69 & 43.69 & - & - & 96.57 & 96.57 & 46.79 & 46.79 & - & - \\
        \textbf{B6\textsubscript{4}} & 98.33 & 98.33 & 69.51 & 69.51 & - & - & 98.30 & 98.30 & 71.40 & 71.40 & - & - \\
        \textbf{HB} & 100 & 100 & 100 & 100 & - & - & 100 & 100 & 100 & 100 & - & -  \\
        \hline 
        \textbf{B\textsubscript{2}} & \textbf{89.55} & 87.35 & 29.80 & 26.89 & 86.00 & 59.08 & 89.25 & 87.20 & 34.78 & 31.72 & 85.93 & 62.18 \\
        \textbf{B\textsubscript{3}} & 89.53 & \textbf{88.60} & 30.94 & 28.16 & \textbf{\underline{86.17}} & 64.49 & \textbf{89.43} & \textbf{88.67} & 35.77 & 32.70 & \textbf{\underline{86.08}} & 66.03 \\
        \textbf{B6\textsubscript{3}} & 86.47 & 84.34 & 19.14 & 17.50 & 82.24 & 53.78 & 86.63 & 84.67 & 23.02 & 21.41 & 82.05 & 56.14 \\
        \textbf{B6\textsubscript{4}} & 88.34 & 87.92 & 25.29 & 22.91 & 81.28 & 53.56 & 88.08 & 87.72 & 29.52 & 27.18 & 81.08 & 57.47 \\
        \textbf{HB} & 88.98 & 88.06 & \textbf{\underline{32.09}} & \textbf{\underline{29.14}} & 83.27 & \textbf{\underline{67.49}} & 88.82 & 88.09 & \textbf{\underline{37.63}} & \textbf{\underline{34.35}} & 83.70 & \textbf{\underline{74.77}} \\
        \hline 
        \textbf{Biaf} & \underline{91.38} & \underline{90.67} & 31.43 & 28.48 & - & - & \underline{91.02} & \underline{90.38} & 37.14 & 34.06 & - & - \\
        \hline 
    \end{tabular}
    \caption{\label{tab:zh-pas-results}Performance in the Chinese-PAS (\textit{zh}\textsubscript{PAS}) dataset. Same notation as in Table \ref{tab:en-dm-results}.}
\end{table*}

\begin{table*}[h]\centering\footnotesize    
    \setlength{\tabcolsep}{2pt}
    \renewcommand{\arraystretch}{1.1}
    \begin{tabular}{|c|cccc|cc|cccc|cc|}
        \cline{2-13}
        \multicolumn{1}{c|}{}& \multicolumn{6}{c|}{\textbf{dev}} & \multicolumn{6}{c|}{\textbf{id}} \\
        \cline{2-13}
        \multicolumn{1}{c|}{}& UF & LF & UM & LM & A & W & UF & LF & UM & LM & A & W \\
        \hline 
        \textbf{B\textsubscript{2}} & 99.98 & 99.98 & 98.46 & 98.46 & - & - & 99.97 & 99.97 & 98.09 & 98.09 & - & - \\
        \textbf{B\textsubscript{3}} & *100 & *100 & 99.78 & 99.78 & - & - & *100 & *100 & 99.85 & 99.85 & - & - \\
        \textbf{B6\textsubscript{3}} & 99.98 & 99.98 & 98.79 & 98.79 & - & - & 99.97 & 99.97 & 98.68 & 98.68 & - & - \\
        \textbf{B6\textsubscript{4}} & 99.99 & 99.99 & 99.45 & 99.45 & - & - & 99.98 & 99.98 & 99.56 & 99.56 & - & - \\
        \textbf{HB} & 100 & 100 & 100 & 100 & - & - & 100 & 100 & 100 & 100 & - & -  \\
        \hline 
        \textbf{B\textsubscript{2}} & 86.73 & \textbf{82.62} & 16.39 & 7.92 & 83.32 & 45.11 & 88.77 & 83.13 & 18.97 & 11.47 & 84.91 & 40.15 \\
        \textbf{B\textsubscript{3}} & 86.88 & 80.84 & 16.28 & 6.93 & 83.34 & 48.70 & 89.01 & 82.75 & 19.12 & 10.15 & 85.14 & 43.88 \\
        \textbf{B6\textsubscript{3}} & 87.37 & 81.38 & 18.26 & 8.36 & 84.15 & 58.07 & 88.89 & 82.72 & 20.59 & 11.62 & \textbf{\underline{85.90}} & 51.46 \\
        \textbf{B6\textsubscript{4}} & \textbf{87.70} & 81.63 & \textbf{\underline{19.47}} & \textbf{\underline{8.58}} & \textbf{\underline{84.45}} & 61.09 & \textbf{89.20} & \textbf{83.17} & \textbf{\underline{22.79}} & \textbf{\underline{13.24}} & 85.80 & 50.29 \\
        \textbf{HB} & 87.37 & 81.29 & 18.70 & 8.36 & 82.87 & \textbf{\underline{65.45}} & 88.90 & 82.54 & 21.32 & 11.47 & 84.03 & \textbf{\underline{54.68}} \\
        \hline 
        \textbf{Biaf} & \underline{89.38} & \underline{84.72} & 16.61 & 6.16 & - & - & \underline{91.24} & \underline{85.26} & 18.24 & 11.03 & - & - \\
        \hline 
    \end{tabular}
    \caption{\label{tab:ar-padt-results}Performance in the Arabic-PADT (\textit{ar}\textsubscript{PADT}) dataset. Same notation as in Table \ref{tab:en-dm-results}.}
\end{table*}

\begin{table*}[h]\centering\footnotesize    
    \setlength{\tabcolsep}{2pt}
    \renewcommand{\arraystretch}{1.1}
    \begin{tabular}{|c|cccc|cc|cccc|cc|}
        \cline{2-13}
        \multicolumn{1}{c|}{}& \multicolumn{6}{c|}{\textbf{dev}} & \multicolumn{6}{c|}{\textbf{id}} \\
        \cline{2-13}
        \multicolumn{1}{c|}{}& UF & LF & UM & LM & A & W & UF & LF & UM & LM & A & W \\
        \hline 
        \textbf{B\textsubscript{2}} & *100 & *100 & 99.91 & 99.91 & - & - & 100 & 100 & 100 & 100 & - & - \\
        \textbf{B\textsubscript{3}} & 100 & 100 & 100 & 100 & - & - & 100 & 100 & 100 & 100 & - & - \\
        \textbf{B6\textsubscript{3}} & 99.98 & 99.98 & 99.10 & 99.10 & - & - & 99.99 & 99.99 & 99.64 & 99.64 & - & - \\
        \textbf{B6\textsubscript{4}} & *100 & *100 & 99.91 & 99.91 & - & - & 99.99 & 99.99 & 99.73 & 99.73 & - & - \\
        \textbf{HB} & 100 & 100 & 100 & 100 & - & - & 100 & 100 & 100 & 100 & - & -  \\
        \hline 
        \textbf{B\textsubscript{2}} & 94.72 & 93.05 & 65.20 & 46.10 & 93.04 & 82.21 & 95.74 & 93.97 & 64.61 & 47.76 & 93.01 & 81.15 \\
        \textbf{B\textsubscript{3}} & 94.76 & 91.56 & 63.86 & 43.86 & 92.85 & 78.27 & 95.72 & 92.89 & 65.05 & 47.94 & 93.04 & 79.65 \\
        \textbf{B6\textsubscript{3}} & 94.86 & \textbf{93.11} & 66.73 & 46.64 & \textbf{\underline{93.69}} & 87.67 & 95.95 & \textbf{94.29} & \textbf{\underline{68.82}} & \textbf{\underline{51.52}} & \textbf{\underline{94.31}} & \textbf{\underline{88.77}} \\
        \textbf{B6\textsubscript{4}} & \textbf{94.95} & 91.83 & \textbf{\underline{67.26}} & 46.55 & 93.66 & 87.46 & \textbf{96.09} & 93.32 & 68.28 & 50.45 & 94.18 & 86.83 \\
        \textbf{HB} & 94.47 & 91.35 & 66.64 & \textbf{47.09} & 92.29 & \textbf{\underline{91.38}} & 95.58 & 92.83 & 67.29 & 50.18 & 92.03 & 88.62 \\
        \hline 
        \textbf{Biaf} & \underline{95.63} & \underline{93.84} & 65.92 & \underline{47.35} & - & - & \underline{96.69} & \underline{94.92} & 66.94 & 48.75 & - & - \\
        \hline 
    \end{tabular}
    \caption{\label{tab:bg-btb-results}Performance in the Bulgarian-BTB (\textit{bg}\textsubscript{BTB}) dataset. Same notation as in Table \ref{tab:en-dm-results}.}
\end{table*}

\begin{table*}[h]\centering\footnotesize    
    \setlength{\tabcolsep}{2pt}
    \renewcommand{\arraystretch}{1.1}
    \begin{tabular}{|c|cccc|cc|cccc|cc|}
        \cline{2-13}
        \multicolumn{1}{c|}{}& \multicolumn{6}{c|}{\textbf{dev}} & \multicolumn{6}{c|}{\textbf{id}} \\
        \cline{2-13}
        \multicolumn{1}{c|}{}& UF & LF & UM & LM & A & W & UF & LF & UM & LM & A & W \\
        \hline 
        \textbf{B\textsubscript{2}} & 99.94 & 99.94 & 98.31 & 98.31 & - & - & 99.93 & 99.93 & 97.94 & 97.94 & - & - \\
        \textbf{B\textsubscript{3}} & 99.99 & 99.99 & 99.63 & 99.63 & - & - & *100 & *100 & 99.87 & 99.87 & - & - \\
        \textbf{B6\textsubscript{3}} & 99.93 & 99.93 & 98.53 & 98.53 & - & - & 99.93 & 99.93 & 98.39 & 98.39 & - & - \\
        \textbf{B6\textsubscript{4}} & 99.98 & 99.98 & 99.34 & 99.34 & - & - & 99.97 & 99.97 & 99.16 & 99.16 & - & - \\
        \textbf{HB} & 100 & 100 & 100 & 100 & - & - & 100 & 100 & 100 & 100 & - & -  \\
        \hline
        \textbf{B\textsubscript{2}} & 92.65 & 89.57 & 53.37 & 41.64 & 88.12 & 61.33 & 93.01 & 90.35 & 53.38 & 42.57 & 88.80 & 61.50 \\
        \textbf{B\textsubscript{3}} & 92.77 & 89.94 & 54.25 & 42.89 & 88.22 & 63.63 & 93.28 & 90.50 & 54.41 & 43.47 & 89.20 & 62.46 \\
        \textbf{B6\textsubscript{3}} & 93.16 & 90.28 & 57.77 & 44.94 & 89.91 & 71.08 & \textbf{93.72} & \textbf{91.03} & \textbf{\underline{58.71}} & \textbf{\underline{46.24}} & \textbf{\underline{91.13}} & 76.32 \\
        \textbf{B6\textsubscript{4}} & \textbf{93.26} & \textbf{90.40} & \textbf{\underline{58.28}} & \textbf{\underline{46.19}} & \textbf{\underline{89.93}} & 73.44 & 93.64 & 90.81 & 58.46 & 45.47 & 90.91 & 78.91 \\
        \textbf{HB} & 92.41 & 89.62 & 56.89 & 44.79 & 88.69 & \textbf{\underline{83.84}} & 92.71 & 90.10 & 56.91 & 45.59 & 89.26 & \textbf{\underline{80.48}} \\
        \hline 
        \textbf{Biaf} & \underline{94.48} & \underline{91.63} & 58.21 & 45.16 & - & - & \underline{95.09} & \underline{92.33} & 58.26 & 45.79 & - & - \\
        \hline 
    \end{tabular}
    \caption{\label{tab:fi-tdt-results}Performance in the Finnish-TDT (\textit{fi}\textsubscript{TDT}) dataset. Same notation as in Table \ref{tab:en-dm-results}.}
\end{table*}

\begin{table*}[h]\centering\footnotesize    
    \setlength{\tabcolsep}{2pt}
    \renewcommand{\arraystretch}{1.1}
    \begin{tabular}{|c|cccc|cc|cccc|cc|}
        \cline{2-13}
        \multicolumn{1}{c|}{}& \multicolumn{6}{c|}{\textbf{dev}} & \multicolumn{6}{c|}{\textbf{id}} \\
        \cline{2-13}
        \multicolumn{1}{c|}{}& UF & LF & UM & LM & A & W & UF & LF & UM & LM & A & W \\
        \hline 
        \textbf{B\textsubscript{2}} & *100 & *100 & 99.76 & 99.76 & - & - & *100 & *100 & 99.56 & 99.56 & - & - \\
        \textbf{B\textsubscript{3}} & 100 & 100 & 100 & 100 & - & - & 100 & 100 & 100 & 100 & - & - \\
        \textbf{B6\textsubscript{3}} & 99.97 & 99.97 & 98.30 & 98.30 & - & - & 99.97 & 99.97 & 98.46 & 98.46 & - & - \\
        \textbf{B6\textsubscript{4}} & 99.99 & 99.99 & 99.27 & 99.27 & - & - & 100 & 100 & 100 & 100 & - & - \\
        \textbf{HB} & 100 & 100 & 100 & 100 & - & - & 100 & 100 & 100 & 100 & - & -  \\
        \hline 
        \textbf{B\textsubscript{2}} & 93.54 & 92.27 & 40.53 & 35.44 & 90.94 & 54.45 & 93.72 & 92.98 & 41.01 & 37.72 & 90.09 & 52.60 \\
        \textbf{B\textsubscript{3}} & 94.01 & 92.15 & 39.81 & 33.01 & 91.04 & 49.89 & 93.87 & 92.47 & 41.23 & 38.16 & 90.67 & 50.14 \\
        \textbf{B6\textsubscript{3}} & \textbf{95.49} & \textbf{94.04} & \textbf{50.97} & \textbf{\underline{42.48}} & \textbf{\underline{93.10}} & 64.42 & 94.90 & \textbf{93.98} & 50.00 & 45.18 & 92.26 & 63.37 \\
        \textbf{B6\textsubscript{4}} & 95.23 & 93.37 & 49.03 & 41.26 & 93.06 & 70.25 & \textbf{95.22} & 93.88 & \textbf{\underline{54.17}} & \textbf{\underline{47.81}} & \textbf{\underline{92.43}} & 70.09 \\
        \textbf{HB} & 94.49 & 92.49 & 48.54 & 40.29 & 91.53 & \textbf{\underline{72.61}} & 94.50 & 92.97 & 51.10 & 44.96 & 89.93 & \textbf{\underline{73.62}} \\
        \hline 
        \textbf{Biaf} & \underline{96.47} & \underline{94.44} & \underline{51.21} & 40.78 & - & - & \underline{96.22} & \underline{94.91} & 52.41 & 47.59 & - & - \\
\hline
    \end{tabular}
    \caption{\label{tab:fr-sequoia-results}Performance in the French-SEQUOIA (\textit{fr}\textsubscript{SEQ.}) dataset.  Same notation as in Table \ref{tab:en-dm-results}.}
\end{table*}

\begin{table*}[h]\centering\footnotesize    
    \setlength{\tabcolsep}{2pt}
    \renewcommand{\arraystretch}{1.1}
    \begin{tabular}{|c|cccc|cc|cccc|cc|}
        \cline{2-13}
        \multicolumn{1}{c|}{}& \multicolumn{6}{c|}{\textbf{dev}} & \multicolumn{6}{c|}{\textbf{id}} \\
        \cline{2-13}
        \multicolumn{1}{c|}{}& UF & LF & UM & LM & A & W & UF & LF & UM & LM & A & W \\
        \hline 
        \textbf{B\textsubscript{2}} & 100 & 100 & 100 & 100 & - & - & *100 & *100 & 99.79 & 99.79 & - & - \\
        \textbf{B\textsubscript{3}} & 100 & 100 & 100 & 100 & - & - & 100 & 100 & 100 & 100 & - & - \\
        \textbf{B6\textsubscript{3}} & 100 & 100 & 100 & 100 & - & - & *100 & *100 & 99.59 & 99.59 & - & - \\
        \textbf{B6\textsubscript{4}} & 100 & 100 & 100 & 100 & - & - & 100 & 100 & 100 & 100 & - & - \\
        \textbf{HB} & 100 & 100 & 100 & 100 & - & - & 100 & 100 & 100 & 100 & - & -  \\
        \hline 
        \textbf{B\textsubscript{2}} & 94.40 & 92.60 & 53.37 & 43.79 & 92.43 & 75.31 & 94.97 & 93.16 & 54.98 & 45.44 & 92.71 & 67.81 \\
        \textbf{B\textsubscript{3}} & 94.36 & 92.40 & 55.14 & 44.50 & 92.57 & 77.09 & 95.53 & 93.47 & 56.22 & 45.02 & 93.21 & 72.97 \\
        \textbf{B6\textsubscript{3}} & 94.62 & 92.74 & 57.27 & 46.63 & 93.06 & 78.68 & 95.46 & 93.51 & 59.96 & 47.93 & 93.93 & 78.00 \\
        \textbf{B6\textsubscript{4}} & \textbf{94.95} & \textbf{93.01} & \textbf{\underline{59.40}} & \textbf{\underline{49.11}} & \textbf{\underline{93.10}} & \textbf{\underline{85.41}} & \textbf{95.58} & 93.52 & \textbf{\underline{60.37}} & 48.76 & \textbf{\underline{93.95}} & 81.19 \\
        \textbf{HB} & 94.50 & 92.69 & 59.40 & 48.40 & 91.93 & 84.78 & 95.55 & \textbf{93.75} & 59.96 & \textbf{\underline{49.38}} & 91.78 & \textbf{\underline{84.99}} \\
        \hline 
        \textbf{Biaf} & \underline{95.40} & \underline{93.35} & 54.26 & 44.33 & - & - & \underline{96.31} & \underline{94.36} & 57.47 & 48.13 & - & - \\
        \hline
    \end{tabular}
    \caption{\label{tab:it-isdt-results}Performance in the Italian-ISDT (\textit{it}\textsubscript{ISDT}) dataset. Same notation as in Table \ref{tab:en-dm-results}.}
\end{table*}

\begin{table*}[h]\centering\footnotesize    
    \setlength{\tabcolsep}{2pt}
    \renewcommand{\arraystretch}{1.1}
    \begin{tabular}{|c|cccc|cc|cccc|cc|}
        \cline{2-13}
        \multicolumn{1}{c|}{}& \multicolumn{6}{c|}{\textbf{dev}} & \multicolumn{6}{c|}{\textbf{id}} \\
        \cline{2-13}
        \multicolumn{1}{c|}{}& UF & LF & UM & LM & A & W & UF & LF & UM & LM & A & W \\
        \hline 
        \textbf{B\textsubscript{2}} & 99.98 & 99.98 & 98.54 & 98.54 & - & - & 99.92 & 99.92 & 97.22 & 97.22 & - & - \\
        \textbf{B\textsubscript{3}} & 100 & 100 & 100 & 100 & - & - & 99.99 & 99.99 & 99.56 & 99.56 & - & - \\
        \textbf{B6\textsubscript{3}} & 99.98 & 99.98 & 98.87 & 98.87 & - & - & 99.97 & 99.97 & 98.68 & 98.68 & - & - \\
        \textbf{B6\textsubscript{4}} & 100 & 100 & 100 & 100 & - & - & 99.99 & 99.99 & 99.71 & 99.71 & - & - \\
        \textbf{HB} & 100 & 100 & 100 & 100 & - & - & 100 & 100 & 100 & 100 & - & -  \\
        \hline 
        \textbf{B\textsubscript{2}} & 86.72 & 84.99 & 26.58 & 18.48 & 82.70 & 39.08 & 85.56 & 83.50 & 29.97 & 18.57 & 79.13 & 41.71 \\
        \textbf{B\textsubscript{3}} & 88.04 & 83.91 & 28.04 & 20.42 & 82.66 & 35.94 & 86.23 & 80.79 & 31.58 & 19.74 & 78.78 & 36.63 \\
        \textbf{B6\textsubscript{3}} & 88.74 & \textbf{86.54} & 32.09 & 21.23 & 85.07 & 57.52 & 87.05 & \textbf{\underline{84.85}} & 36.99 & 21.93 & \textbf{\underline{81.27}} & 59.38 \\
        \textbf{B6\textsubscript{4}} & \textbf{89.24} & 85.18 & \textbf{\underline{34.04}} & \textbf{\underline{22.20}} & \textbf{\underline{85.40}} & \textbf{\underline{63.11}} & \textbf{87.59} & 81.75 & \textbf{\underline{37.57}} & \textbf{\underline{22.37}} & 80.89 & 61.42 \\
        \textbf{HB} & 86.72 & 82.35 & 30.31 & 21.07 & 83.04 & 63.11 & 84.54 & 79.12 & 31.87 & 20.32 & 78.04 & \textbf{\underline{62.30}} \\
        \hline
        \textbf{Biaf} & \underline{91.65} & \underline{87.43} & 30.96 & 21.56 & - & - & \underline{89.63} & 83.82 & 33.33 & 20.47 & - & - \\
        \hline
    \end{tabular}
    \caption{\label{tab:lt-alksnis-results}Performance in the Lithuanian-ALKSNIS (\textit{lt}\textsubscript{ALK.}) dataset. Same notation as in Table \ref{tab:en-dm-results}.}
\end{table*}

\begin{table*}[h]\centering\footnotesize    
    \setlength{\tabcolsep}{2pt}
    \renewcommand{\arraystretch}{1.1}
    \begin{tabular}{|c|cccc|cc|cccc|cc|}
        \cline{2-13}
        \multicolumn{1}{c|}{}& \multicolumn{6}{c|}{\textbf{dev}} & \multicolumn{6}{c|}{\textbf{id}} \\
        \cline{2-13}
        \multicolumn{1}{c|}{}& UF & LF & UM & LM & A & W & UF & LF & UM & LM & A & W \\
        \hline 
        \textbf{B\textsubscript{2}} & 99.95 & 99.95 & 97.96 & 97.96 & - & - & 99.95 & 99.95 & 98.41 & 98.41 & - & - \\
        \textbf{B\textsubscript{3}} & *100 & *100 & 99.82 & 99.82 & - & - & 99.99 & 99.99 & 99.78 & 99.78 & - & - \\
        \textbf{B6\textsubscript{3}} & 99.92 & 99.92 & 97.54 & 97.54 & - & - & 99.94 & 99.94 & 98.03 & 98.03 & - & - \\
        \textbf{B6\textsubscript{4}} & 99.97 & 99.97 & 99.22 & 99.22 & - & - & 99.98 & 99.98 & 99.23 & 99.23 & - & - \\
        \textbf{HB} & 100 & 100 & 100 & 100 & - & - & 100 & 100 & 100 & 100 & - & -  \\
        \hline 
        \textbf{B\textsubscript{2}} & 91.90 & 88.83 & 48.20 & 37.32 & 86.56 & 60.15 & 90.83 & 87.80 & 47.94 & 37.47 & 86.65 & 65.00 \\
        \textbf{B\textsubscript{3}} & 92.12 & 89.01 & 48.74 & 37.56 & 86.42 & 57.79 & 90.47 & 87.12 & 46.19 & 35.27 & 86.01 & 60.67 \\
        \textbf{B6\textsubscript{3}} & 92.09 & 89.13 & 50.60 & 38.40 & \textbf{\underline{88.33}} & 71.54 & 90.93 & 87.78 & 49.20 & 38.29 & \textbf{\underline{88.29}} & 75.01 \\
        \textbf{B6\textsubscript{4}} & \textbf{92.68} & \textbf{89.58} & \textbf{\underline{52.70}} & \textbf{\underline{39.96}} & 88.09 & 71.39 & \textbf{91.51} & \textbf{88.30} & \textbf{\underline{51.89}} & \textbf{\underline{39.99}} & 88.21 & 73.76 \\
        \textbf{HB} & 91.58 & 88.73 & 50.54 & 39.30 & 85.97 & \textbf{\underline{77.99}} & 90.54 & 87.46 & 49.75 & 38.51 & 86.25 & \textbf{\underline{82.57}} \\
        \hline 
        \textbf{Biaf.} & \underline{93.71} & \underline{90.73} & 50.36 & 38.76 & - & - & \underline{93.18} & \underline{89.84} & 51.84 & 39.66 & - & - \\
        \hline
    \end{tabular}
    \caption{\label{tab:lt-alksnis-results}Performance in the Latvian-LVTB (\textit{lv}\textsubscript{LVTB}) dataset. Same notation as in Table \ref{tab:en-dm-results}.}
\end{table*}

\begin{table*}[h]\centering\footnotesize    
    \setlength{\tabcolsep}{2pt}
    \renewcommand{\arraystretch}{1.1}
    \begin{tabular}{|c|cccc|cc|cccc|cc|}
        \cline{2-13}
        \multicolumn{1}{c|}{}& \multicolumn{6}{c|}{\textbf{dev}} & \multicolumn{6}{c|}{\textbf{id}} \\
        \cline{2-13}
        \multicolumn{1}{c|}{}& UF & LF & UM & LM & A & W & UF & LF & UM & LM & A & W \\
        \hline 
        \textbf{B}\textsubscript{2} & 99.99 & 99.99 & 99.54 & 99.54 & - & - & 99.99 & 99.99 & 99.65 & 99.65 & - & - \\
        \textbf{B}\textsubscript{3} & *100 & *100 & 99.98 & 99.98 & - & - & 100 & 100 & 100 & 100 & - & - \\
        \textbf{B6}\textsubscript{3} & *100 & *100 & 99.86 & 99.86 & - & - & *100 & *100 & 99.85 & 99.85 & - & - \\
        \textbf{B6}\textsubscript{4} & *100 & *100 & 99.94 & 99.94 & - & - & *100 & *100 & 99.95 & 99.95 & - & - \\
        \textbf{HB} & 100 & 100 & 100 & 100 & - & - & 100 & 100 & 100 & 100 & - & -  \\
        \hline 
        \textbf{B}\textsubscript{2} & 94.72 & 92.95 & 59.61 & 50.15 & 92.41 & 82.34 & 95.23 & 93.74 & 61.76 & 53.12 & 93.24 & 81.90 \\
        \textbf{B}\textsubscript{3} & 94.84 & 93.11 & 60.10 & 50.88 & 92.38 & 80.83 & 95.15 & 93.64 & 61.52 & 52.47 & 93.22 & 82.18 \\
        \textbf{B6}\textsubscript{3} & 94.81 & 93.06 & 61.47 & 51.50 & 93.10 & 85.44 & 95.30 & 93.86 & 63.21 & 54.01 & \textbf{\underline{93.99}} & 87.52 \\
        \textbf{B6\textsubscript{4}} & \textbf{94.98} & \textbf{93.19} & 62.07 & \textbf{\underline{52.05}} & \textbf{\underline{93.15}} & 88.80 & \textbf{95.39} & \textbf{93.93} & 63.20 & 54.00 & 93.99 & 89.74 \\
        \textbf{HB} & 94.76 & 92.96 & \textbf{\underline{62.20}} & 51.91 & 91.67 & \textbf{\underline{90.45}} & 95.01 & 93.54 & \textbf{\underline{63.32}} & \textbf{\underline{54.24}} & 92.46 & \textbf{\underline{91.81}} \\
        \hline 
        \textbf{Biaf} & \underline{95.59} & \underline{93.81} & 60.09 & 50.65 & - & - & \underline{95.96} & \underline{94.45} & 60.68 & 51.93 & - & - \\
        \hline
    \end{tabular}
    \caption{\label{tab:lt-alksnis-results}Performance in the Russian-SYNTAGRUS (\textit{ru}\textsubscript{SYN.}) dataset. Same notation as in Table \ref{tab:en-dm-results}.}
\end{table*}

\begin{table*}[h]\centering\footnotesize    
    \setlength{\tabcolsep}{2pt}
    \renewcommand{\arraystretch}{1.1}
    \begin{tabular}{|c|cccc|cc|cccc|cc|}
        \cline{2-13}
        \multicolumn{1}{c|}{}& \multicolumn{6}{c|}{\textbf{dev}} & \multicolumn{6}{c|}{\textbf{id}} \\
        \cline{2-13}
        \multicolumn{1}{c|}{}& UF & LF & UM & LM & A & W & UF & LF & UM & LM & A & W \\
        \hline 
        \textbf{B\textsubscript{2}} & 99.99 & 99.99 & 99.53 & 99.53 & - & - & 99.99 & 99.99 & 99.72 & 99.72 & - & - \\
        \textbf{B\textsubscript{3}} & 100 & 100 & 100 & 100 & - & - & 100 & 100 & 100 & 100 & - & - \\
        \textbf{B6\textsubscript{3}} & *100 & *100 & 99.91 & 99.91 & - & - & 99.99 & 99.99 & 99.62 & 99.62 & - & - \\
        \textbf{B6\textsubscript{4}} & 100 & 100 & 100 & 100 & - & - & *100 & *100 & 99.91 & 99.91 & - & - \\
        \textbf{HB} & 100 & 100 & 100 & 100 & - & - & 100 & 100 & 100 & 100 & - & -  \\
        \hline 
        \textbf{B\textsubscript{2}} & 95.13 & 92.10 & 64.25 & 51.32 & 91.67 & 74.42 & 94.92 & 92.08 & 63.71 & 50.90 & 90.96 & 72.80 \\
        \textbf{B\textsubscript{3}} & 95.39 & 92.20 & 63.96 & 50.00 & 91.91 & 73.41 & 94.69 & 91.83 & 61.64 & 49.29 & 90.90 & 74.11 \\
        \textbf{B6\textsubscript{3}} & 94.92 & 92.06 & 66.70 & 52.08 & 92.90 & 79.81 & \textbf{95.43} & \textbf{92.51} & 66.92 & 52.87 & 92.63 & 80.16 \\
        \textbf{B6\textsubscript{4}} & \textbf{95.68} & \textbf{92.81} & \textbf{\underline{69.25}} & 53.40 & \textbf{\underline{93.24}} & 84.79 & 95.30 & 92.41 & \textbf{67.48} & 53.16 & \textbf{\underline{92.66}} & 83.75 \\
        \textbf{HB} & 94.87 & 92.31 & 66.89 & \textbf{54.15} & 91.95 & \textbf{\underline{88.42}} & 94.62 & 92.11 & 66.35 & \textbf{\underline{54.85}} & 91.25 & \textbf{\underline{85.19}} \\
        \hline
        \textbf{Biaf} & \underline{96.78} & \underline{93.97} & 68.77 & \underline{54.43} & - & - & \underline{96.81} & \underline{93.99} & \underline{68.61} & 54.38 & - & - \\
        \hline
    \end{tabular}
    \caption{\label{tab:sk-snk-results}Performance in the Slovak-SNK (\textit{sk}\textsubscript{SNK}) dataset. Same notation as in Table \ref{tab:en-dm-results}.}
\end{table*}

\begin{table*}[h]\centering\footnotesize    
    \setlength{\tabcolsep}{2pt}
    \renewcommand{\arraystretch}{1.1}
    \begin{tabular}{|c|cccc|cc|cccc|cc|}
        \cline{2-13}
        \multicolumn{1}{c|}{}& \multicolumn{6}{c|}{\textbf{dev}} & \multicolumn{6}{c|}{\textbf{id}} \\
        \cline{2-13}
        \multicolumn{1}{c|}{}& UF & LF & UM & LM & A & W & UF & LF & UM & LM & A & W \\
        \hline 
        \textbf{B\textsubscript{2}} & *100 & *100 & 99.80 & 99.80 & - & - & *100 & *100 & 99.84 & 99.84 & - & - \\
        \textbf{B\textsubscript{3}} & 100 & 100 & 100 & 100 & - & - & 100 & 100 & 100 & 100 & - & - \\
        \textbf{B6\textsubscript{3}} & 99.97 & 99.97 & 98.21 & 98.21 & - & - & 99.98 & 99.98 & 99.34 & 99.34 & - & - \\
        \textbf{B6\textsubscript{4}} & 99.99 & 99.99 & 99.40 & 99.40 & - & - & *100 & *100 & 99.92 & 99.92 & - & - \\
        \textbf{HB} & 100 & 100 & 100 & 100 & - & - & 100 & 100 & 100 & 100 & - & -  \\
        \hline 
        \textbf{B\textsubscript{2}} & 89.60 & 87.01 & 34.13 & 28.57 & 85.49 & 53.70 & 92.42 & 89.90 & 45.69 & 38.64 & 88.05 & 60.85 \\
        \textbf{B\textsubscript{3}} & 89.26 & 86.41 & 34.33 & 26.98 & 85.39 & 52.42 & 92.01 & 89.50 & 44.63 & 38.23 & 88.33 & 59.94 \\
        \textbf{B6\textsubscript{3}} & \textbf{91.21} & \textbf{88.54} & 40.67 & 32.94 & \textbf{\underline{87.76}} & 57.28 & \textbf{93.32} & \textbf{91.12} & \textbf{\underline{53.49}} & \textbf{\underline{45.28}} & \textbf{\underline{90.77}} & 69.53 \\
        \textbf{B6\textsubscript{4}} & 90.93 & 87.92 & \textbf{\underline{41.07}} & \textbf{\underline{33.13}} & 87.41 & 56.62 & 93.14 & 90.97 & 50.78 & 43.23 & 90.53 & 68.07 \\
        \textbf{HB} & 89.84 & 87.06 & 40.08 & 31.75 & 84.73 & \textbf{\underline{69.08}} & 92.16 & 89.87 & 49.71 & 42.00 & 88.30 & \textbf{\underline{76.20}} \\
        \hline 
        \textbf{Biaf} & \underline{93.04} & \underline{89.97} & 37.10 & 28.77 & - & - & \underline{94.59} & \underline{92.05} & 48.89 & 41.43 & - & - \\

        \hline
    \end{tabular}
    \caption{\label{tab:sv-talbanken-results}Performance in the Swedish-TALBANKEN (\textit{sv}\textsubscript{TAL.}) dataset. Same notation as in Table \ref{tab:en-dm-results}.}
\end{table*}

\begin{table*}[h]\centering\footnotesize    
    \setlength{\tabcolsep}{2pt}
    \renewcommand{\arraystretch}{1.1}
    \begin{tabular}{|c|cccc|cc|cccc|cc|}
        \cline{2-13}
        \multicolumn{1}{c|}{}& \multicolumn{6}{c|}{\textbf{dev}} & \multicolumn{6}{c|}{\textbf{id}} \\
        \cline{2-13}
        \multicolumn{1}{c|}{}& UF & LF & UM & LM & A & W & UF & LF & UM & LM & A & W \\
        \hline 
        \textbf{B\textsubscript{2}} & 100 & 100 & 100 & 100 & - & - & 100 & 100 & 100 & 100 & - & - \\
        \textbf{B\textsubscript{3}} & 100 & 100 & 100 & 100 & - & - & 100 & 100 & 100 & 100 & - & - \\
        \textbf{B6\textsubscript{3}} & 100 & 100 & 100 & 100 & - & - & 100 & 100 & 100 & 100 & - & - \\
        \textbf{B6\textsubscript{4}} & 100 & 100 & 100 & 100 & - & - & 100 & 100 & 100 & 100 & - & - \\
        \textbf{HB} & 100 & 100 & 100 & 100 & - & - & 100 & 100 & 100 & 100 & - & -  \\
        \hline 
        \textbf{B\textsubscript{2}} & 78.22 & 67.10 & 11.25 & \textbf{\underline{6.25}} & 72.09 & 27.43 & 74.15 & 61.89 & 7.50 & 1.67 & 74.77 & 44.15 \\
        \textbf{B\textsubscript{3}} & 78.22 & 67.10 & 11.25 & 6.25 & 72.09 & 27.43 & 74.15 & 61.89 & 7.50 & 1.67 & 74.77 & 44.15 \\
        \textbf{B6\textsubscript{3}} & \textbf{\underline{79.69}} & \textbf{\underline{69.31}} & \textbf{\underline{15.00}} & 6.25 & 76.81 & \textbf{\underline{63.29}} & \textbf{\underline{78.18}} & \textbf{\underline{66.08}} & \textbf{\underline{11.67}} & 2.50 & 77.01 & 55.27 \\
        \textbf{B6\textsubscript{4}} & 79.69 & 69.31 & 15.00 & 6.25 & 76.81 & 63.29 & 78.18 & 66.08 & 11.67 & 2.50 & 77.01 & 55.27 \\
        \textbf{HB} & 76.87 & 66.36 & 15.00 & 6.25 & \textbf{\underline{77.18}} & 62.86 & 75.33 & 63.67 & 11.67 & \textbf{\underline{3.33}} & \textbf{\underline{77.49}} & \textbf{\underline{74.77}} \\
        \hline
        \textbf{Biaf} & 79.43 & 67.80 & 8.75 & 5.00 & - & - & 76.69 & 64.83 & 9.17 & 2.50 & - & - \\
        \hline
    \end{tabular}
    \caption{\label{tab:ta-ttb-results}Performance in the Tamil-TTB (\textit{ta}\textsubscript{TTB}) dataset. Same notation as in Table \ref{tab:en-dm-results}.}
\end{table*}

\begin{table*}[h]\centering\footnotesize    
    \setlength{\tabcolsep}{2pt}
    \renewcommand{\arraystretch}{1.1}
    \begin{tabular}{|c|cccc|cc|cccc|cc|}
        \cline{2-13}
        \multicolumn{1}{c|}{}& \multicolumn{6}{c|}{\textbf{dev}} & \multicolumn{6}{c|}{\textbf{id}} \\
        \cline{2-13}
        \multicolumn{1}{c|}{}& UF & LF & UM & LM & A & W & UF & LF & UM & LM & A & W \\
        \hline 
        \textbf{B\textsubscript{2}} & 99.99 & 99.99 & 99.40 & 99.40 & - & - & 99.99 & 99.99 & 99.10 & 99.10 & - & - \\
        \textbf{B\textsubscript{3}} & 100 & 100 & 100 & 100 & - & - & 100 & 100 & 100 & 100 & - & - \\
        \textbf{B6\textsubscript{3}} & 99.99 & 99.99 & 99.55 & 99.55 & - & - & 99.99 & 99.99 & 99.10 & 99.10 & - & - \\
        \textbf{B6\textsubscript{4}} & *100 & *100 & 99.85 & 99.85 & - & - & *100 & *100 & 99.78 & 99.78 & - & - \\
        \textbf{HB} & 100 & 100 & 100 & 100 & - & - & 100 & 100 & 100 & 100 & - & -  \\
        \hline 
        \textbf{B\textsubscript{2}} & 92.54 & 90.53 & 39.58 & 33.78 & 87.49 & 55.52 & 92.39 & 89.94 & 40.92 & 34.30 & 86.84 & 49.92 \\
        \textbf{B\textsubscript{3}} & 92.25 & 89.88 & 38.99 & 32.59 & 87.60 & 54.14 & 92.27 & 89.96 & 39.01 & 32.74 & 86.61 & 52.20 \\
        \textbf{B6\textsubscript{3}} & \textbf{93.20} & \textbf{91.02} & 43.90 & 36.01 & \textbf{\underline{89.10}} & 66.24 & 92.85 & 90.38 & 44.62 & 37.33 & 88.44 & 63.98 \\
        \textbf{B6\textsubscript{4}} & 93.07 & 90.88 & \textbf{\underline{45.68}} & \textbf{\underline{36.61}} & 88.90 & 68.32 & \textbf{93.03} & \textbf{90.59} & \textbf{\underline{46.30}} & \textbf{37.89} & \textbf{\underline{88.58}} & 61.34 \\
        \textbf{HB} & 91.97 & 89.84 & 42.86 & 35.86 & 86.50 & \textbf{\underline{75.81}} & 91.81 & 89.50 & 43.72 & 37.44 & 85.37 & \textbf{\underline{67.84}} \\
        \hline 
        \textbf{Biaf} & \underline{94.71} & \underline{92.28} & 43.15 & 34.38 & - & - & \underline{94.50} & \underline{92.17} & 44.96 & \underline{38.34} & - & - \\
        \hline
    \end{tabular}
    \caption{\label{tab:uk-iu-results}Performance in the Ukrainian-IU (\textit{uk}\textsubscript{IU}) dataset. Same notation as in Table \ref{tab:en-dm-results}.}
\end{table*}

\end{document}